%% file: 0_main_TPAMI.tex
\documentclass[10pt,journal,compsoc]{IEEEtran}
\usepackage[utf8]{inputenc}
\usepackage{amsmath,amsfonts}
\usepackage{hyperref}
\usepackage{amssymb}
\usepackage{algorithm}
\usepackage{bbding} 
\usepackage{algpseudocode}
\usepackage{color}
\usepackage{multirow}
\usepackage{array}
\usepackage{tikz}
\usetikzlibrary{shadows}
\usetikzlibrary{trees,positioning,shapes,shadows,arrows}
\usepackage[edges]{forest}
\definecolor{hiddendraw}{RGB}{205, 44, 36}
\usepackage[caption=false,font=normalsize,labelfont=sf,textfont=sf]{subfig}
\usepackage{textcomp}
\usepackage{stfloats}
\usepackage{url}
\usepackage{verbatim}
\usepackage{graphicx}
\usepackage{cite}
\newcommand{\tabincell}[2]{\begin{tabular}{@{}#1@{}}#2\end{tabular}}
\newcommand{\sunrgbd}{SUN RGB-D}
\newcommand{\scannet}{ScanNetV2}

\ifCLASSINFOpdf
\else
\fi
\begin{document}
\title{Transformers in 3D Point Clouds: A Survey}

\author{
Dening Lu, 
Qian Xie,
Mingqiang Wei,~\IEEEmembership{Senior Member,~IEEE},
Kyle (Yilin) Gao,~\IEEEmembership{Student Member,~IEEE},
Linlin Xu,~\IEEEmembership{Member,~IEEE,}
and Jonathan Li,~\IEEEmembership{Senior Member,~IEEE}
\IEEEcompsocitemizethanks{\IEEEcompsocthanksitem Dening Lu and Qian Xie contribute equally to this work and should be considered co-first authors.
\IEEEcompsocthanksitem Corresponding authors: Linlin Xu; Jonathan Li.% <-this % stops an unwanted space
\IEEEcompsocthanksitem Dening Lu, Kyle Gao, Linlin Xu, and Jonathan Li are with the Department of Systems Design Engineering, University of Waterloo, Waterloo, Ontario N2L 3G1, Canada (e-mail: {d62lu, y56gao, l44xu, junli}@uwaterloo.ca).% <-this % stops an unwanted space
\IEEEcompsocthanksitem Mingqiang Wei is with the Shenzhen Research Institute, Nanjing University of Aeronautics and Astronautics, Shenzhen 518038, China (e-mail: mingqiang.wei@gmail.com).% <-this % stops an unwanted space
\IEEEcompsocthanksitem Jonathan Li is also with the Department of Geography and Environmental Management, University of Waterloo, Waterloo, Ontario N2L 3G1, Canada.% <-this % stops an unwanted space
\IEEEcompsocthanksitem Qian Xie is with the Department of Computer Science, University of Oxford, Oxford OX1 3QD, U.K. (e-mail: qian.xie@cs.ox.ac.uk).}% <-this % stops an unwanted space
% \thanks{Dening Lu and Qian Xie contribute equally to this work and should be considered co-first authors.}% <-this % stops a space
% \thanks{Corresponding authors: Linlin Xu; Jonathan Li.}% <-this % stops a space
% \thanks{Dening Lu, Linlin Xu are with the Department of Systems Design Engineering, University of Waterloo, Waterloo, ON N2L 3G1,  Canada (e-mail: d62lu@uwaterloo.ca; linlinxu618@gmail.com).}
% \thanks{Jonathan Li is with the Department of Geography and  Environmental Management, University of Waterloo, Waterloo, ON N2L 3G1,  Canada (e-mail: junli@uwaterloo.ca).}
% \thanks{Qian Xie is with the the Department of Computer Science, University of Oxford, Oxford OX1 3QD, U.K. (e-mail: qian.xie@cs.ox.ac.uk).}
}

\IEEEtitleabstractindextext{%
\begin{abstract}
Transformers have been at the heart of the Natural Language Processing (NLP) and Computer Vision (CV) revolutions. The significant success in NLP and CV inspired exploring the use of Transformers in point cloud processing. However, how do Transformers cope with the irregularity and unordered nature of point clouds? How suitable are Transformers for different 3D representations (e.g., point- or voxel-based)? How competent are Transformers for various 3D processing tasks? As of now, there is still no systematic survey of the research on these issues. For the first time, we provided a comprehensive overview of increasingly popular Transformers for 3D point cloud analysis. We start by introducing the theory of the Transformer architecture and reviewing its applications in 2D/3D fields. Then, we present three different taxonomies (i.e., implementation-, data representation-, and task-based), which can classify current Transformer-based methods from multiple perspectives. Furthermore, we present the results of an investigation of the variants and improvements of the self-attention mechanism in 3D. To demonstrate the superiority of Transformers in point cloud analysis, we present comprehensive comparisons of various Transformer-based methods for classification, segmentation, and object detection. Finally, we suggest three potential research directions, providing benefit references for the development of 3D Transformers.

\end{abstract}

\begin{IEEEkeywords}
Transformer, point cloud analysis, self-attention mechanism, deep neural networks, 3D vision.
\end{IEEEkeywords}}

\maketitle

\IEEEdisplaynontitleabstractindextext

% \fi
\IEEEpeerreviewmaketitle

%-------------------------
\input{1_Intro}

\input{2_Transformer_Implementation}
\input{3_data_representation}

\input{4_Task}

\input{5_Variants}

\input{6_Comparison_and_Analysis}
\input{7_conclusion_future}
%-------------------------

\bibliographystyle{IEEEtran}
\bibliography{mybibfile}

\vfill

\end{document}

%% file: 1_Intro.tex
\IEEEraisesectionheading{\section{Introduction}
\label{sec:1}}

\begin{figure*}[htbp]
\centering
\includegraphics[width=0.8\linewidth]{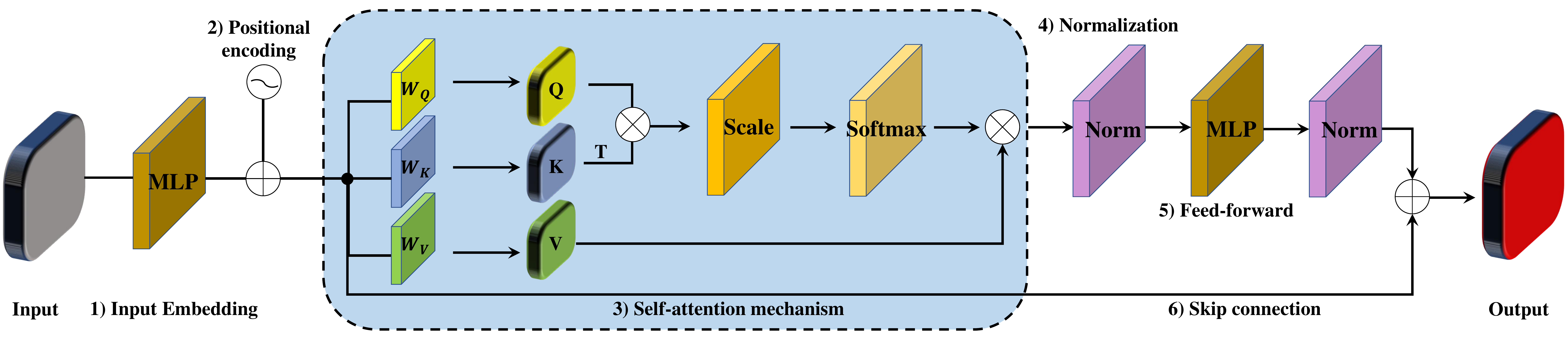}
\caption{Illustration of the Transformer encoder architecture.}
\label{fig:transformer_structure}
\end{figure*}

\IEEEPARstart{T}{ransformers}, in encoder and/or decoder configurations, are now the dominant neural architecture in NLP. In view of the impressive ability to model long-range dependencies, they have been successfully adapted to the field of CV~\cite{han2022survey,li2022contextual,xiao2022image} for autonomous driving, visual computing, intelligent monitoring, and industrial inspection. A standard Transformer encoder generally consists of six main components (Fig. \ref{fig:transformer_structure}): 1) input (word) embedding; 2) positional encoding; 3) self-attention mechanism; 4) normalization; 5) feed-forward operation; and 6) skip connection.
As for the Transformer decoder, it is typically designed to mirror the Transformer encoder, except it additionally takes as input latent features from the Transformer-encoder. 
However, for 3D point cloud applications, decoders can be specifically designed (i.e. not be a pure Transformer) for dense prediction tasks such as part segmentation and semantic segmentation in 3D point cloud analysis. 
Researchers in 3D computer vision often adopt PointNet++ \cite{qi2017pointnet++} or convolutional backbones with Transformer blocks incorporated therein. 

To describe in more detail, let $P = \left\{p_{1}, p_{2}, p_{3}, ..., p_{N} \right\} \in R^{N \times D}$ be an input point cloud. $D$ is the feature dimension of the input point. Typically in literature, ``$D$ equals to three" means only the 3D coordinate of each point is taken as input, while ``$D$ equals to six" means both the 3D coordinate and normal vector are taken as input. The details of the aforementioned encoder components are as follows.
% \begin{itemize}
    % \item 

Firstly, for the input embedding, $P$ is projected to a high-dimension feature space which can facilitate subsequent learning. This can be achieved by using a Multi-Layer Perception (MLP) or other feature extraction backbone networks like PointNet \cite{qi2017pointnet}. We denote the embedded feature map as $X \in R^{N \times C}$.
% \item 
Secondly, the positional encoding is used to either capture the geometrical information, or the relative ordering of input tokens/points if relevant. Note that the Transformer is order-agnostic without this step, which is not an issue for point clouds, since they are naturally unordered. Nonetheless, the frequency-based positional encoding can be used by mapping spatial coordinates with sine and cosine functions \cite{vaswani2017attention}. Moreover, there also exist learned position encoding schemes with a trainable parameter matrix $B$ \cite{zhao2021point, lu20223dctn}, which are more adaptive to different input data. These positional encodings of spatial coordinates have shown to benefit the learning of features at finer scales \cite{tancik2020fourier}. 
% \item 
Thirdly, the core component of the Transformer encoder is the self-attention mechanism. If a sine/cosine-based positional encoding is used, it is typically added to the embedded feature map $X$. This feature map is then projected to three different feature spaces using three learnable weight matrices $W_{Q} \in R^{C \times C_{Q}}, W_{k} \in R^{C \times C_{K}}, W_{V} \in R^{C \times C}$, where typically $ C_{K}$ equals to  $C_{Q}$. In this way, $Query$, $Key$, and $Value$ matrices can be formulated as:
\begin{equation}
\begin{aligned}
\left\{\begin{matrix}
 Query &= X  W_{Q},\\
 Key &= X  W_{K},\\
 Value &= X  W_{V}.
\end{matrix}\right.
\end{aligned}
\end{equation}
Given the $Query$, $Key$, and $Value$ matrices, an attention map is formulated as:
\begin{equation}
    Attentionmap = Softmax(\frac{Q K^{T}}{\sqrt{C_{K}}}),
\label{eq:1}
\end{equation}
where $Q,K,V$ denote the $Query$, $Key$, and $Value$ matrices respectively.
The attention map of size of $N \times N$ measures the similarity of any two input points. It is also called the similarity matrix. Then the attention map and the matrix $Value$  are multiplied to generate the new feature map $F$, of the same size as $X$. Each feature vector in $F$ is obtained by computing a weighted sum of all input features. It is therefore able to establish connections with all input features. When inputs are global, this process allows for the Transformer to easily learn global features. Therefore, compared with convolutional neural networks (CNNs), Transformers are better at long-range dependency modeling. Additionally, compared with MLPs, Transformers also have two significant advantages. One is that the attention map in the Transformer is dynamic depending on the input during the inference, which is more adaptive than MLPs with fixed weight matrices. Another is that the self-attention mechanism is permutation-equivariant, while for MLPs, the order of input and output is encoded in the weight matrix.
% \item 
Fourthly, a normalization layer is placed before and after the feed-forward layer, performing standardization and normalization on feature maps. There are two kinds of normalization methods used in this layer: LayerNormalization and BatchNormalization. The former is commonly used in NLP, while the latter is commonly used in CV like 2D or 3D data processing. 
% \item 
Fifthly, a feed-forward layer is added to enhance the representation of  attention features. Generally, it consists of two fully-connection layers with a RELU function. 
% \item 
Finally, a skip connection is used between the input and output of the self-attention module. There have been many self-attention variants using various skip connection forms \cite{feng2020point, xie2018attentional, guo2021pct}, which we present in more details in Sec. \ref{sec:5}. 
% \end{itemize}

Note that there are also some 3D Transformers that are not exactly comprised of these six components. 
For example, early 3D Transformer networks like Point Attention (P-A) \cite{feng2020point} and Attentional ShapeContextNet \cite{xie2018attentional} did not have the positional encoding module. They focused on applying the self-attention mechanism to 3D point clouds. Point Cloud Transformer (PCT) \cite{guo2021pct} proposed a neighbor embedding mechanism achieved by EdgeConv \cite{wang2019dynamic}. This mechanism incorporates the positional encoding into the input embedding module. 
Since the self-attention mechanism is the core component of Transformers, we also classify methods mainly utilizing the self-attention mechanism for point cloud processing into the 3D Transformer family for this survey.

Recently, Transformer models have been introduced to image processing widely, and achieved impressive results for various tasks, such as image segmentation \cite{wang2021max}, object detection \cite{carion2020end} and tracking \cite{chen2021transformer}. 
Vision Transformer (ViT) \cite{dosovitskiy2020image} first proposed a pure Transformer network for image classification. It achieved excellent performance compared with the state-of-the-art convolutional networks.
Based on ViT, there were numerous Transformer variants proposed for image classification \cite{wu2020visual, wang2021pyramid, wu2021cvt, liu2021swin}, segmentation \cite{xie2021segformer, cheng2021per, wang2021end}, object detection \cite{carion2020end, yao2021efficient, wang2021pyramid, yang2021focal}, and other vision tasks. Moreover, various innovations on Transformer-based architectures were proposed, such as convolutions+Transformers \cite{wu2021cvt},
multi-scale Transformers \cite{liu2021swin},  and self-supervised Transformers \cite{chen2021empirical}. 
There also have been several surveys \cite{liu2021survey, khan2021transformers, han2022survey, xu2022multimodal} proposed to categorize all involved 2D Transformers into multiple groups. The taxonomies they used were algorithm architecture-based taxonomy and task-based taxonomy.

\begin{figure}[htbp]
\centering
\includegraphics[width=\linewidth]{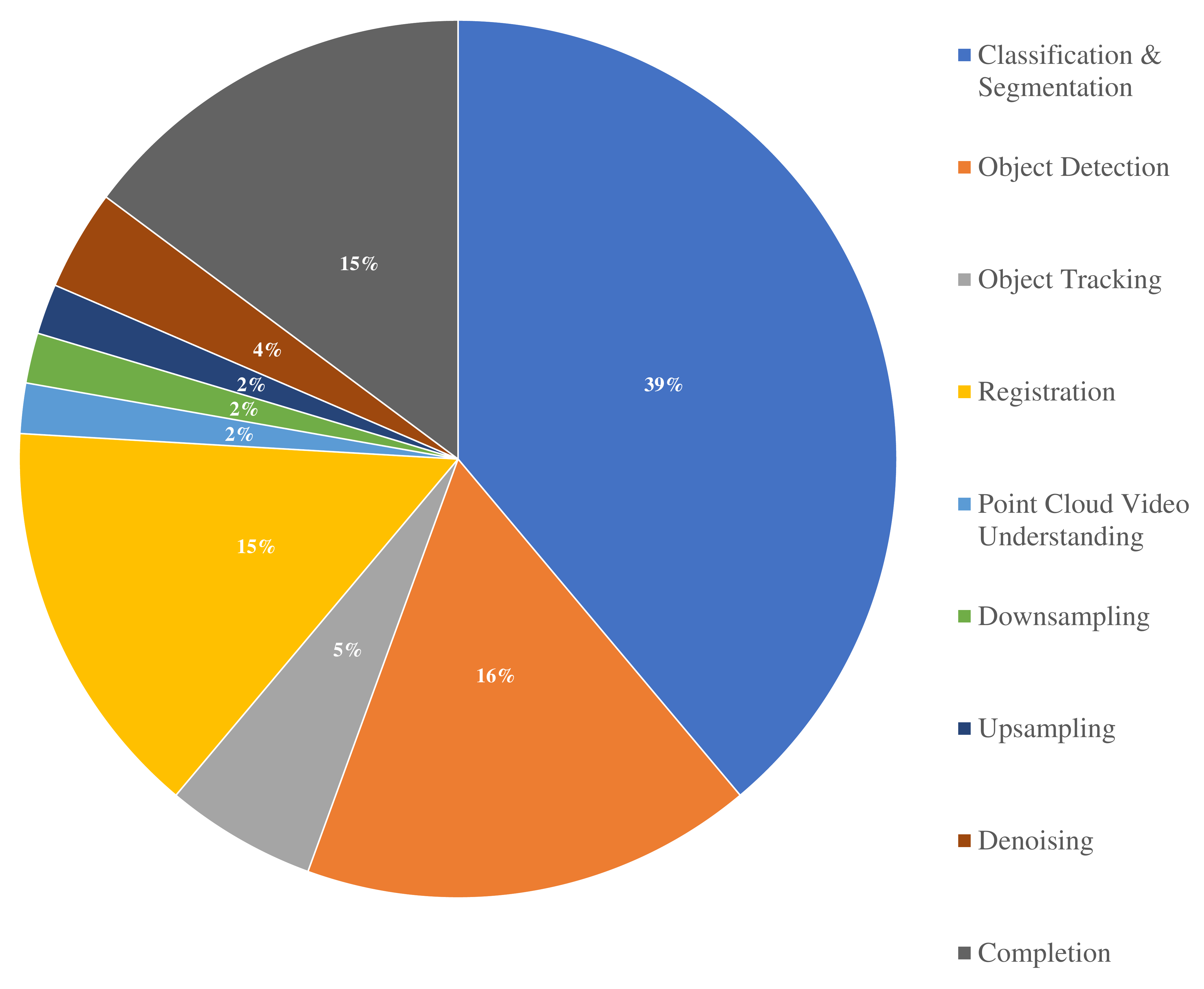}
\caption{Applications of Transformers in 3D point clouds.}
\label{fig:task_proportion}
\end{figure}

\begin{figure*}[htbp]
 \centering
\begin{forest}
  forked edges,
  for tree={
    grow=east,
    reversed=true,%increase counter-clockwise
    anchor=base west,
    parent anchor=east,
    child anchor=west,
    base=left,
    font=\small,
    rectangle,
    draw={hiddendraw, line width=0.6pt},
    rounded corners,align=left,
    minimum width=2.5em,
    edge={black, line width=0.55pt},
    l+=1.9mm,
    s sep=7pt,
    inner xsep=7pt,
    inner ysep=8pt,
    ver/.style={rotate=90, rectangle, draw=none, rounded corners=3mm, fill=red, drop shadow, text centered,  text=white, child anchor=north, parent anchor=south, anchor=center, font=\fontsize{10}{10}\selectfont,},
    level2/.style={rectangle, draw=none, fill=orange,  drop shadow,
    text centered, anchor=west, text=white, font=\fontsize{8}{8}\selectfont, text width = 7em},
    level3/.style={rectangle, draw=none, fill=brown,  drop shadow, fill opacity=0.8,text centered, anchor=west, text=white, font=\fontsize{8}{8}\selectfont, text width = 2.7cm, align=center},
    level3_1/.style={rectangle, draw=none, fill=brown,  drop shadow, fill opacity=0.8, text centered, anchor=west, text=white, font=\fontsize{8}{8}\selectfont, text width = 3.2cm, align=center},
    level4/.style={rectangle, draw=red, text centered, anchor=west, text=black, font=\fontsize{8}{8}\selectfont, align=center, text width = 2.9cm},
    level5/.style={rectangle, draw=red, text centered, anchor=west, text=black, font=\fontsize{8}{8}\selectfont, align=center, text width = 14.7em},
    level5_1/.style={rectangle, draw=red, text centered, anchor=west, text=black, font=\fontsize{8}{8}\selectfont, align=center, text width = 26.1em},
  },
  where level=1{text width=5em,font=\scriptsize,align=center}{},
  where level=2{text width=6em,font=\tiny,}{},
  where level=3{text width=6em,font=\tiny}{},%yshift=0.26pt
  where level=4{text width=5em,font=\tiny}{},%yshift=0.26pt
  where level=5{font=\tiny}{},%yshift=0.26pt
  [3D Transformers, ver
    [Implementations, level2
        [Operating Scale, level3
            [Global Transformers, level4
                [PCT \cite{guo2021pct}{,}
                3CROSSNet \cite{han20223crossnet}{,}\\
                PointASNL \cite{yan2020pointasnl}{,}
                Point-BERT \cite{yu2021pointbert}, level5]
            ]
            [Local Transformers, level4
                [PT \cite{zhao2021point}{,}
                LFT-Net \cite{gao2022lft}{,}
                Pointformer \cite{pan20213d}, level5]
            ]
        ]
        [Operating Space, level3
            [Point-wise Transformers, level4
                [PT \cite{zhao2021point}{,}
                3DCTN \cite{lu20223dctn}{,}
                PCT \cite{guo2021pct}{,}\\
                TD-Net \cite{xu2022tdnet}{,}
                3DMedPT \cite{yu20213d}, level5]
            ]
            [Channel-wise \\ Transformers, level4
                [PU-Transformer \cite{qiu2021pu}{,}\\
                Dual Transformer \cite{han2021dual}{,}\\
                CAA \cite{qiu2022geometric}{,}
                ACE-Transformer\cite{xu2021adaptive}, level5]
            ]
        ]
        [Efficient Transformers, level3
            [Centroid-Transformer \cite{wu2021centroid}{,}
            LighTN \cite{wang2022lightn}{,}
            GSA \cite{yang2019modeling}, level5_1]
            ]
    ]
    [Data Representations, level2
        [Voxel-based \\Transformers, level3
            [PVT \cite{zhang2021pvt}{,} 
            VoTr~\cite{mao2021voxel}{,} 
            VoxSeT~\cite{he2022voxset}{,}
            SVT-Net~\cite{fan2021svt}{,} 
            EPT~\cite{park2022efficient}, level5_1]
        ]
        [Point-based\\ Transformers, level3
            [Uniform  Scale, level4
                [PCT~\cite{guo2021pct}{,} 
                Point-BERT \cite{yu2021pointbert}{,}\\
                PCTMA-Net \cite{lin2021pctma}, level5]
            ]
            [Multi Scale, level4
                [PT~\cite{zhao2021point}{,}
                3DCTN \cite{lu20223dctn}{,}\\
                Stratified Transformer \cite{lai2022stratified}{,}
                PPT \cite{hui2021pyramid}, level5]
            ]
        ]
    ]
    [3D Tasks, level2
        [High-level Task, level3
            [Classification \& \\ Segmentation, level4
                [Point Transformer~\cite{zhao2021point}{,}
                PCT~\cite{guo2021pct}{,}\\
                DT-Net~\cite{han2021dual}{,}
                PVT~\cite{zhang2021pvt}, level5]
            ]
            [Object Detection, level4
                [MLCVNet~\cite{xie2020mlcvnet}{,}
                Group-free~\cite{liu2021group}{,}\\
                3DETR~\cite{misra2021end}{,}
                VoTr~\cite{mao2021voxel}, level5]
            ]
            [Object Tracking, level4
                [LTTR~\cite{cui20213d}{,}
                PTTR~\cite{zhou2021pttr}{,}
                PTT~\cite{jiayao2022real}, level5]
            ]
            [Registration, level4
                [DCP~\cite{wang2019deep}{,}
                Storm~\cite{wang2022storm}{,} 
                Stickypillars~\cite{fischer2021stickypillars}{,}\\
                RGM~\cite{fu2021robust}{,}
                DIT~\cite{chen2021full}, level5]
            ]
            [Point Cloud \\ Video Understanding, level4
                [P4Transformer~\cite{fan2021point}, level5]
            ]
        ]
        [Low-level Task, level3
            [Downsampling, level4
                [LighTN~\cite{wang2022lightn}, level5]
            ]
            [Upsampling, level4
                [PU-Transformer~\cite{qiu2021pu}, level5]
            ]
            [Denosing, level4
                [TDNet~\cite{xu2022tdnet}{,}
                Gao et al.~\cite{gao2022reflective}, level5]
            ]
            [Completion, level4
                [PCTMA-Net~\cite{lin2021pctma}{,}
                PointTr~\cite{yu2021pointr}{,} \\SnowflakeNet~\cite{xiang2021snowflakenet}{,} 
                VQDIF~\cite{yan2022shapeformer}, level5]
            ]
        ]
    ]
]
\end{forest}
\caption{Taxonomies of 3D Transformers.}
\label{fig:taxonomy}
\end{figure*}
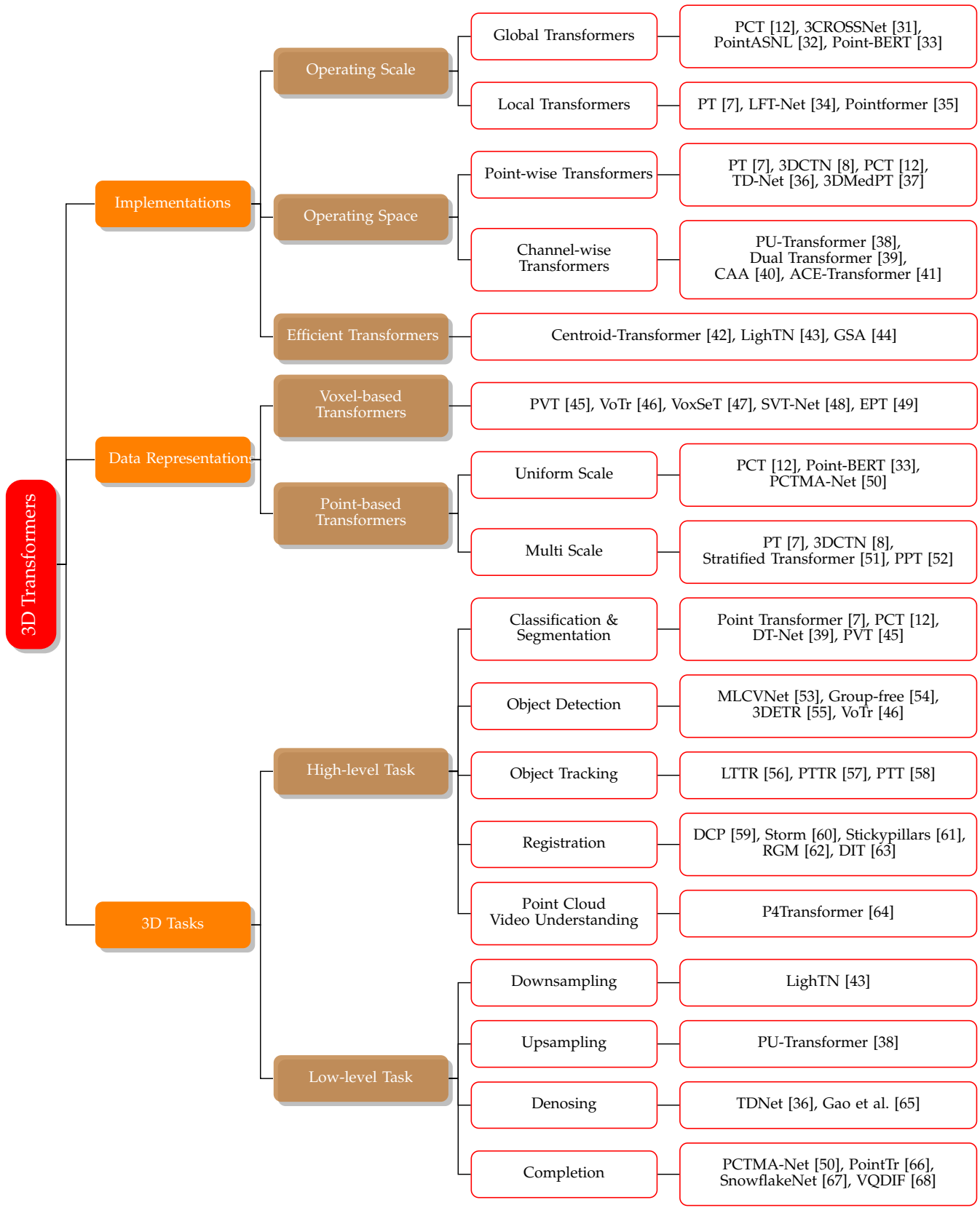

% \tikzset{
%     my node/.style={
%         draw=red,
%         inner color=gray!5,
%         outer color=gray!10,
%         thick,
%         minimum width=1cm,
% %           rounded corners=3,
%         text height=1.5ex,
%         text depth=0ex,
%         font=\sffamily,
%         drop shadow,
%     }
% }
% \begin{figure*}
%     \centering
% \begin{forest}
%     for tree={%
%         my node,
%         l sep+=5pt,
%         grow'=east,
%         edge={gray, thick},
%         parent anchor=east,
%         child anchor=west,
%         if n children=0{tier=last}{},
%         edge path={
%             \noexpand\path [draw, \forestoption{edge}] (!u.parent anchor) -- +(10pt,0) |- (.child anchor)\forestoption{edge label};
%         },
%         if={isodd(n_children())}{
%             for children={
%                 if={equal(n,(n_children("!u")+1)/2)}{calign with current}{}
%             }
%         }{}
%     }
%     [A
%     [b
%     [l]]
%     [c
%     [e[m]][f[n]][g[o]][h[p]]]
%     [d
%     [j[r]][k[s]]]
%     ]
% \end{forest}
%     \caption{Three taxonomies for 3D Transformer classification}
%     \label{fig:taxonomy}
% \end{figure*}

Due to the remarkable global feature learning ability and permutation-equivariant operations, Transformer architectures are intrinsically suited for point cloud processing and analysis. 
A number of 3D Transformer backbones were proposed (see  Fig. \ref{fig:task_proportion}) for point cloud classification \& segmentation\cite{guo2021pct, sheng2021improving, zhao2021point, gao2022lft, yu20213d, wei2022spatial}, detection \cite{pan20213d, liu2021group}, tracking \cite{cui20213d,zhou2021pttr,jiayao2022real}, registration \cite{wang2019deep,wang2022storm,fischer2021stickypillars,fu2021robust,chen2021full,qin2022geometric,yew2022regtr}, completion~\cite{yu2021pointr,xiang2021snowflakenet,chen2021transsc,lin2021pctma,yan2022shapeformer,liu2022point}, to name a few. 
% The satisfactory performance achieved by these networks has proven that the 3D Transformer is beneficial to point cloud feature extraction and further scene understanding. 
Moreover, 3D Transformer networks have also been used for various practical applications, such as structure monitoring \cite{zhou2022sewer}, medical data analysis \cite{yu20213d}, and autonomous driving \cite{prakash2021multi, yuan2021temporal}. 
Therefore, it is necessary to conduct a systematic survey for 3D Transformers. 
Recently, several 3D Transformer/Attention-related reviews have been published.
For instance, Khan et al. \cite{khan2021transformers} reviewed the vision Transformers according to the architecture- and task-based taxonomies. 
However, it mainly focused on Transformers on 2D image analysis, and only provided a brief introduction to 3D Transformer networks.
Qiu et al. \cite{qiu2021investigating}  introduced several variants of the 3D self-attention mechanism, and conducted a detailed comparison and analysis for them on SUN RGBD \cite{song2015sun} and ScanNetV2 datasets \cite{dai2017scannet}. 
However, a comprehensive survey of Transformer models in 3D point clouds has not been conducted so far, which we hope to provide with this paper.
%This survey provides a comprehensive investigation of 3D Transformers, starting from the aforementioned existing review works. 

We designed three different taxonomies which are shown in Fig. \ref{fig:taxonomy}: 1) implementation-based taxonomy; 2) data representation-based taxonomy; 3) task-based taxonomy. In this way, we were able to classify and analyze Transformer networks from multiple perspectives. We note that these taxonomies are not mutually exclusive. Taking Point Transformer (PT) \cite{zhao2021point} as an example: 1) in terms of the Transformer implementation, it belongs to the local Transformer category, operated in the local neighborhood of the target point cloud; 2) in terms of the data representation, it belongs to the multi-scale point-based Transformer category, extracting the geometrical and semantic features hierarchically; 3) in terms of the 3D task, it is designed for point cloud classification and segmentation. Additionally, we also conducted an investigation of different self-attention variants in 3D point cloud processing. We expect this classification to provide helpful references for the development of Transformer-based networks.

The major contributions of this survey can be summarised as follows:
\begin{itemize}
    \item This is the first effort, to the best of our knowledge, that focused on comprehensively covering Transformers in point clouds for 3D vision tasks.
    \item This work investigates a series of self-attention variants in point cloud analysis. It introduced novel self-attention mechanisms aiming to improve the performance and efficiency of 3D Transformers.
    \item This work provides comparisons and analyses of Transformer-based methods on several 3D vision tasks, including 3D shape classification and 3D shape/semantic segmentation, and 3D object detection on several public benchmarks. 
    \item This work introduces the readers to the SOTA methods as well as to the recent progress of Transformers-based methods for point cloud processing.
\end{itemize}

The core of this paper is organized into six sections not counting the Introduction. Sec. \ref{sec:2}, \ref{sec:3}, and \ref{sec:4} introduce the three different taxonomies for 3D Transformer classification. Sec. \ref{sec:5} reviews different self-attention variants proposed in literature to improve the performance of Transformers.
Sec. \ref{sec:6} provides a comparison and analysis of surveyed 3D Transformer networks. Lastly, Sec. \ref{sec:7} summarizes our survey work, and points out three potential future directions for 3D Transformers.

%% file: 2_Transformer_Implementation.tex
\section{Transformer Implementation}
\label{sec:2}

In this section, we broadly categorize Transformers in 3D point clouds from multiple perspectives. Firstly, in terms of the operating scale, 3D Transformers can be divided into two parts: Global Transformers and Local Transformers (Sec. \ref{subsec:2.1}). The operating scale represents the scope of the algorithm with respect to the point cloud, such as the global domain or the local domain. Secondly, in terms of the operating space, 3D Transformers can be divided into Point-wise Transformers and Channel-wise Transformers (Sec. \ref{subsec:2.2}). The operating scale represents the dimension in which the algorithm is operated, such as the spatial dimension or the channel dimension. Lastly, we provide a review of efficient Transformer networks designed for computational footprint reduction (Sec.\ref{subsec:2.3}).

\subsection{Operating Scale}
\label{subsec:2.1}

According to the operating scale, 3D Transformers can be divided into two parts: Global Transformers and Local Transformers. The former denotes that Transformer blocks are applied to all the input points for global feature extraction, while the latter denotes that Transformer blocks are applied in a local patch for local feature extraction.

\subsubsection{Global Transformers}

There are many existing works \cite{xie2018attentional, feng2020point, guo2021pct, hui2021pyramid, gao2021multi, qiu2021pu, yu20213d, yu2021pointbert, han20223crossnet, lu20223dctn} focusing on global Transformer. For a global Transformer block, each new output feature in $F$ can establish connections with all input features $X$. It is both equivariant with respect to permutations of the input and capable of learning the global context features\cite{guo2021pct}.

Following PointNet \cite{qi2017pointnet}, PCT, as a pure global Transformer network, was proposed in \cite{guo2021pct}. Taking the 3D coordinates as input $P$, PCT first proposed a neighbor-embedding architecture to map the point cloud into a high-dimensional feature space. This operation can also incorporate local information into the embedded features. Then these features were fed into four stacked global Transformer blocks to learn semantic information. The global features were finally extracted by a global Max and Average (MA) pooling for classification and segmentation.
% , as shown in Fig. \ref{fig:pct1}. 
Moreover, PCT's improved self-attention module, named Offset-Attention, was inspired by the Laplacian matrix in Graph convolution networks \cite{bruna2013spectral}. We detailed the structure of the Offset-Attention module in Sec.~\ref{subsec:point-wise variant}. It is able to sharpen the attention weights and reduce the influence of noise. The state-of-the-art performance of PCT on various tasks proved that Transformers are suitable for 3D point cloud processing.

% \begin{figure}[htbp]
% \centering
% \includegraphics[width=\linewidth]{fig/Implementation/pct1.pdf}
% \caption{Overview of PCT (from \cite{guo2021pct}).}
% \label{fig:pct1}
% \end{figure}

In contrast to the single scale of PCT, a Cross-Level Cross-Scale Cross-Attention Transformer network was proposed in \cite{han20223crossnet}, named 3CROSSNet. Firstly, it performed Farthest Point Sampling (FPS) algorithm \cite{qi2017pointnet++} on the raw input point cloud to obtain three point subsets with different resolutions. Secondly, it utilized stacked multiple shared Multi-Layer Perception (MLP) modules to extract local features for each sampling point. Thirdly, it applied Transformer blocks to each point subset for global feature extraction. Finally, the Cross-Level Cross-Attention (CLCA) module and Cross-Scale Cross-Attention (CSCA) module were proposed to build connections between different-resolution point subsets and different-level features for long-range inter- and intra-level dependencies modeling.

A BERT-style pre-training strategy for 3D global Transformers was proposed in \cite{yu2021pointbert}, which adapted BERT \cite{devlin2018bert} to 3D point cloud processing. 
% As shown in Fig. \ref{fig:cls-bert}, t
Taking the local patches as input, it first utilized the mini-PointNet \cite{qi2017pointnet} for the input embedding, following ViT \cite{dosovitskiy2020image}. Then it used a point cloud Tokenizer with a discrete Variational AutoEncoder (dVAE) \cite{rolfe2016discrete}, to convert the embedded points into discrete point tokens for pre-training. 
The Tokenizer network was adapted from DGCNN \cite{wang2019dynamic} which produced meaningful local information aggregating, and was learned through dVAE-based point cloud reconstruction. During pre-training, the point embeddings with some masked tokens were fed into the Transformer encoder. Supervised by the point tokens generated by the Tokenizer, the encoder can be trained to recover the corresponding tokens of the masked locations. The authors have conducted comprehensive experiments to show that the BERT-style pre-training strategy is able to improve the performance of the pure Transformer in point cloud classification and segmentation.

\subsubsection{Local Transformers}

In contrast to global Transformers, local Transformers \cite{zhao2021point, wu2021centroid, pan20213d, gao2022lft, wang2022local, liu2022group} aim to achieve feature aggregation in the local patch instead of the entire point cloud. 

% \begin{figure}[htbp]
% \centering
% \includegraphics[width=\linewidth]{fig/Implementation/pt1.pdf}
% \caption{Overview of PT (from \cite{zhao2021point}).}
% \label{fig:pointtransformer}
% \end{figure}

PT~\cite{zhao2021point} adopted the PointNet++ \cite{qi2017pointnet++} hierarchical architecture for point cloud classification and segmentation. 
It focused on local patch processing, and replaced the shared MLP modules in PointNet++ with local Transformer blocks. 
% As shown in Fig. \ref{fig:pointtransformer}, 
PT had five local Transformer blocks operated on progressively downsampled point sets. Each block was applied on K-Nearest Neighbor (KNN) neighborhoods of sample points. Specifically, the self-attention operator that PT used was the vector attention \cite{zhao2020exploring} instead of the scalar attention. The former has been proven to be more effective for point cloud processing, since it supports channel-wise attention weight assignment, as opposed to assigning one single weight to a whole feature vector. Please refer to Sec. \ref{sec:7} an overview of vector attention.

% \begin{figure}[htbp]
% \centering
% \includegraphics[width=\linewidth]{fig/classification/Point_BERT.pdf}
% \caption{Illustration of point-BERT (from \cite{yu2021pointbert}).}
% \label{fig:cls-bert}
% \end{figure}

% \textcolor{red}{Similarly, Local Feature Transformer Network (LFT-Net) \cite{gao2022lft} also attempted to increase the expressiveness of local fine-grained features. It consisted of four stacked local Transformer and Trans-pool blocks, so the local features could be continuously aggregated into the global features. There are two main differences between LFT-Net and PT. The first one is that LFT-Net proposed a trans-pooling model, instead of commonly-used symmetry functions like max/mean/sum pooling. This model was able to alleviate the feature discarding. 
% Another is that LFT-Net applied multi focal-loss instead of the standard cross-entropy loss function. This loss can address the problems of class imbalance and weak learning ability for complex regions in the semantic segmentation task. It differs from the the standard cross-entropy loss by a class-based weight term and a decay coefficient, which are able to balance the data distribution, and enhance the impact of the low-accuracy class on the loss.}

Pointformer was proposed in  \cite{pan20213d} to combine the local and global features both extracted by Transformer blocks for 3D object detection. 
It had three kinds of main blocks: a Local Transformer (LT) block, a Global Transformer (GT) block and a Local-Global Transformer (LGT) block. 
Firstly, the LT block applied the dense self-attention operation in the neighborhood of each centroid point generated by FPS \cite{qi2017pointnet++}.
Secondly, taking the whole point cloud as input, the GT block aimed to learn global context-aware features via the self-attention mechanism. 
Lastly, the LGT block adopted a multi-scale cross-attention module, to build connections between local features from the LT and global features from the GT. 
Specifically, the LGT block took the output of the LT as $query$, and the output of the GT as $key$ and $value$ to conduct the self-attention operation. As such, all centroid points could be utilized to integrate global information, which led to effective global feature learning.

% \begin{figure*}[htbp]
% \centering
% \includegraphics[width=0.95\linewidth]{fig/Implementation/Stratified_Transformer.pdf}
% \caption{Architecture of Stratified Transformer framework (from \cite{lai2022stratified}).}
% \label{fig:stratified}
% \end{figure*}

Inspired by Swin Transformer \cite{liu2021swin}, Stratified Transformer \cite{lai2022stratified} was proposed for 3D point cloud segmentation.
It split the point cloud into a group of non-overlapping cubic windows via 3D voxelization and performed the local Transformer operation in each window. 
% Fig. \ref{fig:stratified}(a) shows the encoder-decoder architecture of the Stratified Transformer framework. 
% As shown in Fig. \ref{fig:stratified}(b), t
Stratified Transformer was an encoder-decoder architecture. Its encoder was a hierarchical structure consisting of multiple stages, where each stage had two successive Transformer blocks. The former block utilized a Stratified Self-Attention (SSA) to capture the long- and short-range dependencies. The latter block utilized a Shifted SSA to further strengthen the connections between different independent windows, following Swin Transformer \cite{liu2021swin}. 
Specifically, to solve the issue that the local Transformer is weak in capturing global information, SSA generated the dense local $key$ points and sparse distant $key$ points for each $query$ point. 
The former was generated in the window that the $query$ point belonged to, while the latter was generated in a larger window by downsampling the entire input point cloud. 
With this, the receptive field of the $query$ point was not limited in the local window, allowing SSA to capture global information. 
Additionally, Stratified Transformer performed a KPConv \cite{thomas2019kpconv} embedding in the first stage to extract the local geometric information of the input point cloud. 
This operation was proven to be effective by their ablation experiments.

\subsection{Operating Space}
\label{subsec:2.2}

According to the operating space, 3D Transformers can be divided into two categories: Point-wise Transformers and Channel-wise Transformers. 
The former measures the similarity among input points, while the latter distributes attention weights along channels \cite{qiu2022geometric}. 
Generally, according to Eq. \ref{eq:1}, the attention maps of these two kinds of Transformers can be expressed as:
\begin{equation}
\begin{aligned}
    Point-wise \; Attn &= Softmax(\frac{Q  K^{T}}{\sqrt{C_{K}}}),\\
    Channel-wise \; Attn &= Softmax(\frac{Q^{T}  K}{\sqrt{C_{K}}}),
\end{aligned}
\label{eq:2}
\end{equation}
where the size of $Point-wise \; Attn$ is $N \times N$, while the size of $Channel-wise \; Attn$ is $C_{K} \times C_{K}$.
 
\subsubsection{Point-wise Transformers}
\label{subsec:2.2.1}
Point-wise Transformers aim to investigate the spatial correlation among points and formulate the output feature map as a weighted sum of all input features. 
% learn the long-range context-dependent representation. I.e., it spatially describes the long-range contextual dependencies \cite{han2021dual}. The output feature map of point-wise Transformers can be formulated as

Since global Transformers and local Transformers in Sec. \ref{subsec:2.1} are distinguished by the spatial operating scale, i.e., the whole point cloud or a local patch, all aforementioned methods \cite{xie2018attentional, feng2020point, guo2021pct, hui2021pyramid, zhao2021point, gao2021multi, wu2021centroid, pan20213d, qiu2021pu, yu20213d, yu2021pointbert, han20223crossnet, gao2022lft, lu20223dctn, lai2022stratified} in Sec. \ref{subsec:2.1} can be considered as point-wise Transformers. 

Point-wise Transformers are also widely applied to other tasks. Xu et al. \cite{xu2022tdnet} proposed an encoder-decoder Transformer network (TD-Net) for point cloud denoising. The encoder consisted of a coordinate-based input embedding module, an adaptive sampling module and four stacked point-wise self-attention modules. The outputs of the four self-attention modules were concatenated together as input of the decoder.
Additionally, TD-Net used the adaptive sampling approach which can automatically learn the offset of each sampling point generated by FPS \cite{qi2017pointnet++}. This operation allows the sampling points closer to the underlying surface.
The decoder was applied to construct the underlying manifold according to the extracted high-level features. And finally, a clean point cloud can be reconstructed by manifold sampling.

3D Medical Point Transformer (3DMedPT) \cite{yu20213d} was proposed for medical point cloud analysis. Specifically, it included a hierarchical point-wise Transformer for classification and a uniform-scale point-wise Transformer for segmentation. 
3DMedPT introduced the convolution operation to the point-wise Transformer block. 
It added a local feature extraction module achieved using DGCNN \cite{wang2019dynamic} before each Transformer block. 
Point Attention Network (P-A) \cite{feng2020point} and Pyramid Point Cloud Transformer (PPT) \cite{hui2021pyramid} also had similar structures. 
Considering insufficient training sample processing in the medical domain, 3DMedPT proposed a special module named Multi-Graph Reasoning (MGR), to enrich the feature representations.

\subsubsection{Channel-wise Transformers}
In contrast to point-wise Transformers, the channel-wise Transformers \cite{qiu2021pu, qiu2022geometric, han2021dual, xu2021adaptive, sheng2021improving} focus on measuring the similarity of different feature channels. 
They are able to improve the context information modeling by highlighting the role of interaction across different channels \cite{han2021dual}. 

% \textcolor{red}{Dual Transformer Network (DT-Net) was proposed in \cite{han2021dual} for point cloud analysis, which applied the point-wise Transformer and channel-wise Transformer operations simultaneously. After the point-wise and channel-wise features were extracted respectively, they were summed element-wise to improve feature representation. Ablation studies in \cite{han2021dual} demonstrated that DT-Net with both point-wise and channel-wise Transformer modules achieved the best results, compared to ablation models with only point-wise or only channel-wise Transformers.}

Qiu et al. \cite{qiu2022geometric} proposed a back-projection module for local feature capturing, leveraging an idea of error-correcting feedback structure. They designed a Channel-wise Affinity Attention (CAA) module for better feature representations.
% as shown in Fig. \ref{fig:caa}. 
Specifically, the CAA module consisted of two blocks: a Compact Channel-wise Comparator (CCC) block and a Channel Affinity Estimator (CAE) block. The CCC block could generate the similarity matrix in the channel space. The CAE block further calculated an affinity matrix, in which an element with a higher attention value represented a lower similarity of the corresponding two channels. This operation can sharpen the attention weights and avoid aggregating similar/redundant information. As such, each channel of the output feature had sufficient intersection with other distinct ones, which has been proven to be beneficial to the final results. We also detail the CAA structure in Sec. \ref{sec:5}.

% \begin{figure}[htbp]
% \centering
% \includegraphics[width=\linewidth]{fig/Implementation/CAA.pdf}
% \caption{\textcolor{red}{Illustration of the CAA module (from \cite{qiu2022geometric}).}}
% \label{fig:caa}
% \end{figure}

Instead of only using the feature channels, Transformer-Conv proposed in \cite{xu2021adaptive} combined both coordinate and feature channels to design a novel channel-wise Transformer. Specifically, 
% as shown in Fig. \ref{fig:trans_conv}, 
the $Query$ matrix was generated directly by the coordinate information without any linear transformation, while the $Key$ matrix was generated by feature channels with MLP. Then the attention matrix was calculated by element-wise multiplication rather than dot product. As such, the attention matrix was able to represent the relationship between the coordinate channels and feature channels of each point. 
Since all features came from the coordinate space, an element with a higher value in the attention matrix tended to indicate that the corresponding feature channel was more faithful to the coordinate space. 
After that, the $Value$ matrix is created by projecting the $Key$ matrix into a latent space using an MLP.
Then the new feature map, called the response matrix, can be obtained by element-wise multiplication between the $Value$ matrix and attention matrix. The response matrix consisted of the weighted feature channels of all input points. Finally, the output features were generated by applying a channel max-pooling operation to the response matrix. The max-pooling played a screening role and it could select the most important channels, i.e., the channels which suited the coordinate space best. This process was proven effective for point cloud analysis by the authors' ablation experiments.

% \begin{figure*}[htbp]
% \centering
% \includegraphics[width=\linewidth]{fig/Implementation/trans-conv_coding.pdf}
% \caption{Flow chart of Transformer-Conv (from \cite{xu2021adaptive}).}
% \label{fig:trans_conv}
% \end{figure*}

\subsection{Efficient Transformers}
\label{subsec:2.3}

Despite achieving great success in point cloud processing, standard Transformers tend to incur high computational footprint and memory consumption because of massive linear operations. Given $N$ input points, the computation and memory complexities of the standard self-attention module are quadratic on $N$, i.e., $O(N_{2})$. This is the key drawback when applying Transformers on large-scale point cloud datasets. 

Recently, there were several 3D Transformers researching on improving the self-attention module for higher computational efficiency. 
For instance, Centroid Transformer \cite{wu2021centroid} took $N$ point features as input while outputting a smaller number $M$ of point features. As such, the key information in the input point cloud can be summarized by a smaller number of outputs (called centroids). Specifically, it first constructed $M$ centroids from $N$ input points, by optimizing a general “soft K-means” objective function. Then it used the $M$ centroids and $N$ input points to generate the $Query$ and $Key$ matrices respectively. The size of the attention map was reduced from $N \times N$ to $M \times N$, so the computational cost of the self-attention was reduced from $O(N_{2})$ to $O(NM)$. To further save the computational cost, the authors applied a KNN approximation. This operation essentially converted the global Transformer to a local Transformer. In this case, the similarity matrix was generated by measuring the relationships among each $query$ feature vector and its $K$ neighbor $key$ vectors, instead of $N$ vectors. So the computational cost can be further reduced to $O(NK)$. Similarly, PatchFormer~\cite{cheng2021patchformer} was also proposed to reduce the size of the attention map. It first split the raw point cloud into $M$ patches, followed by aggregating the local feature in each patch. The significant difference between the two aforementioned models is that PatchFormer used $M$ aggregated local features to generate $Key$ matrix, while Centroid Transformer used $M$ centroids to generate $Query$ matrix. As such, the computational cost of the self-attention in PatchFormer can also be reduced to $O(NM)$. 

Light-weight Transformer Network (LighTN) \cite{wang2022lightn} was proposed to reduce the computational cost in a different way. 
LighTN aimed to simplify the main components in the standard Transformer, maintaining the superior performance of Transformers while increasing efficiency. Firstly, it removed the positional encoding block because the input 3D coordinates already contain positional information and can be considered as a substitute for the positional encoding. This eliminated the overhead of positional encoding itself. Secondly, it utilized a small-size shared linear layer as the input embedding layer. The dimensions of embedded features were reduced by half compared to the computationally-saving neighbor embedding setting in \cite{guo2021pct}. With this, the computational costs of the input embedding can be reduced. Thirdly, it presented a single head self-correlation layer as the self-attention module. The projection matrices of $W_{Q}$, $W_{K}$, and $W_{V}$ were removed, to reduce learnable parameters for high efficiency. Since the attention map was generated only by the input self-correlation parameters, the self-attention module was also named the self-correlation module, which can be formulated as:
\begin{equation}
\label{self-correlation}
\begin{aligned}
SA(X) &= FC_{out}(C(X)),\\
C(X) &= softmax(\frac{X  X^{T}}{\sqrt{C}})  X,
\end{aligned}
\end{equation}
where $SA(\ast)$ represents the self-attention block, $FC_{out}$ represents the linear transformation, $softmax(\ast)$ is the activation function, and $C$ is the input feature dimension declared in Eq. \ref{eq:1}.
% \, \, \, \, $SA(\ast)$ represents the self-attention block, 
% \, \, \, \, $FC_{out}$ represents the linear transformation, 
% \, \, \, \, $softmax(\ast)$ is the activation function,
% \, \, \, \, $C$ is the input feature dimension declared in Eq. \ref{eq:1}.
Lastly, the authors built three linear layers (a standard FFN block generally has two linear layers) in the Feed-Forward Network (FFN) and used the expand-reduce strategy \cite{mehta2020delight} in the middle layer. So the negative impact caused by the decreasing learnable parameters in the self-correlation layer can be mitigated. Similarly, Group Shuffle Attention (GSA) proposed in \cite{yang2019modeling} also simplified the self-attention mechanism in its Transformer network. It integrated the shared projection weight matrix and non-linearity activation function into the self-attention mechanism (please see Sec. \ref{subsec:point-wise variant} for a detailed description of GSA).

%% file: 3_data_representation.tex
\section{Data Representation}
\label{sec:3}
There are several forms of 3D data representation, such as points and voxels, both of which can be used as the input of 3D Transformers.
Since points can be represented by or transformed into voxels, several voxel-based approaches can also be performed on point clouds, so as to 3D Transformers.
According to different input formats, we divided 3D Transformers into Voxel-based Transformers and Point-based Transformers.

% \begin{figure}[htbp]
% \centering
% \includegraphics[width=0.8\linewidth]{fig/voxelization.pdf}
% \caption{Illustration of 3D point cloud voxelization (from \cite{bello2020deep}).}
% \label{fig:voxel}
% \end{figure}

\subsection{Voxel-based Transformers}

Unlike images, 3D point clouds are generally unstructured, and cannot be directly processed by traditional convolution operators.
However, 3D point clouds can be easily converted into 3D voxels, which are structured like images. 
% as illustrated in Fig.~\ref{fig:voxel}.
Thus, some Transformer-based works ~\cite{mao2021voxel, zhang2021pvt, lai2022stratified, he2022voxset} explored transforming 3D point clouds into voxel-based representation.
The most general voxelization approach can be described as follows~\cite{xu2021voxel}: The bounding box of a point cloud is first regularly divided into 3D cuboids via rasterization.
Voxels containing points are retained, generating the voxel representation of point clouds.

Inspired by the efficiency of sparse convolution on voxel data \cite{graham20183d,choy20194d}, Mao et al.~\cite{mao2021voxel} first proposed the Voxel Transformer (VoTr) backbone for 3D object detection.
They presented the Submanifold Voxel module and the Sparse Voxel module to extract features from non-empty and empty voxels respectively.
In both two modules, the Local Attention and Dilated Attention operations were implemented, on the basis of the Multi-head Self-Attention mechanism (MSA), to maintain low computational consumption for numerous voxels.
The proposed VoTr can be integrated into most voxel-based 3D detectors.
To tackle the computational issue of Transformers as voxel-based outdoor 3D detectors, Voxel Set Transformer (VoxSeT)~\cite{he2022voxset} was proposed to detect outdoor objects in a set-to-set fashion.
Based on the low-rank characteristic of the self-attention matrix, a Voxel-based Self-Attention (VSA) module was designed by assigning a set of trainable ``latent codes" to each voxel, which was inspired by the induced set attention blocks in Set Transformer~\cite{lee2019set}. 

Inspired by the effectiveness of voxel-based representation on large-scale point clouds, voxel-based Transformers can also be applied to large-scale point cloud processing.
For instance, Fan et al.~\cite{fan2021svt} presented Super light-weight Sparse Voxel Transformer (SVT-Net) for large scale place recognition.
They designed an Atom-based Sparse Voxel Transformer (ASVT) and a Cluster-based Sparse Voxel Transformer (CSVT). The former was used to encode short-range local relations, while the latter was used to learn long-range contextual relations.
Park et al.~\cite{park2022efficient} proposed Efficient Point Transformer (EPT) for large-scale 3D scene understanding from point clouds.
To relieve the problem of geometric information loss during voxelization, they introduced the center-aware voxelization and devoxelization operations. On this basis, Efficient Self-Attention (ESA) layers were employed to extract voxel features.
Their center-aware voxelization preserved positional information of points in voxels.

\subsection{Point-based Transformers}

Since voxels are of regular format and points are not, the transformation to voxels would lead to geometric information loss to some extent \cite{qi2017pointnet,qi2017pointnet++}. On the other hand, since the point cloud is the original representation, it contains the complete geometric information of the data.
Thus, most Transformer-based point cloud processing frameworks fall into the category of point-based Transformer.
% The mathematical input is a matrix whose rows and columns represent points and the corresponding 3D spatial coordinates.
Their architectures are usually classified in two main groups: uniform-scale architecture~\cite{guo2021pct,chen2021full,lin2021pctma,gao2021multi,yu2021pointbert} and multi-scale architecture~\cite{lu20223dctn, hui2021pyramid, zhao2021point,han2021dual,yu20213d,lai2022stratified}. 
% Besides, some point-based transformers are also combined with convolutions, as several works~\cite{graham2021levit,wu2021cvt,yuan2021incorporating} in 2D Transformers.
\subsubsection{Uniform Scale}

Uniform-scale architectures usually keep the scale of the point features constant during data processing. The number of output features of each module is consistent with the number of input features.
The most representative work is PCT~\cite{guo2021pct}, which was discussed in Sec. \ref{subsec:2.1}.
After the input embedding stage, four global Transformer blocks of PCT were directly stacked together to refine point features. There was no hierarchical feature aggregation operation, which facilitated the decoder design for dense prediction tasks like point cloud segmentation. Feeding all input points into the Transformer block is beneficial to global feature learning. 
However, uniform-scale Transformers tend to be weak in extracting the local features due to the lack of local neighborhoods. 
Additionally, processing the whole point cloud directly would lead to a high computation footprint and memory consumption.

\subsubsection{Multi Scale}

Multi-scale Transformers refer to those with progressive point sampling strategies during feature extraction, also called hierarchical Transformers.
PT~\cite{zhao2021point} was the pioneering design that introduced the multi-scale structure to a pure Transformer network.
% As shown in Fig.~\ref{fig:pointtransformer}, 
The Transformer layers in PT were applied to progressively (sub)sampled point sets.
On one hand, sampling operations could accelerate the computation of the whole network by reducing the parameters of the Transformer.
On the other hand, these hierarchical structures usually came with KNN-based local feature aggregation operations.
This local feature aggregation was beneficial to the tasks that need fine semantic perception, such as segmentation and completion. And the highly aggregated local features at the last layer of the network could be taken as the global features, which could be used for point cloud classification. Additionally, there also exist many multi-scale Transformer networks \cite{lu20223dctn, yu20213d, hui2021pyramid, lai2022stratified} that utilized EdgeConv \cite{wang2019dynamic} or KPconv \cite{thomas2019kpconv} for local feature extraction and utilized Transformers for global feature extraction. With this, they are able to combine the strong local modeling ability of convolutions and the remarkable global feature learning ability of Transformers for better semantic feature representation.

% Convolutions have achieved great success for a long time in the fields of 2D and 3D data processing. As the development of Transformer in 3D point cloud processing, the combination of Transformer and convolutions has attracted  a great deal of attention. DGCNN \cite{wang2019dynamic} proposed an efficient way to extract the local feature of the point cloud, named ``Edge Convolution", which has been widely used in 3D point cloud deep learning frameworks. Therefore, there are also many works exploring to combine the strong local modeling ability of Edge Convolution and remarkable global feature learning ability of Transformer for 3D point cloud processing.

% In this section, we divide the involved works into two groups: Hybird Transformer with CNN and pure Transformer.

% \subsubsection{Hybird Transformer with CNN.}

% xxx...
% 只选择那些上次我们看的结合了DGCNN的算法，之前章节应该都介绍过，这里就简单讲一下每篇文章是怎么结合的DGCNN。

%另外，我想，不要这个pure了，将这个hybrid和前面的multi scale, uniform 并列起来，之前的一篇survey “Transformers in Vision: A Survey” 就是怎么干的 ，里面的hybrid Transformer的具体写法也可以参考一下。这样应该就好写一点了。

% \subsubsection{Pure Transformer.}

% xxx...

%% file: 4_Task.tex
\section{3D Tasks}
\label{sec:4}

Similar to image processing \cite{khan2021transformers}, 3D point cloud-related tasks can also be divided into two main groups: high-level and low-level tasks.
High-level tasks involve semantic analysis, which focuses on translating 3D point clouds to information that people can understand.
Low-level tasks, such as denoising and completion, focus on exploring fundamental geometric information. They are not directly related to human semantic understanding but can indirectly contribute to high-level tasks.

% \begin{figure*}[htbp]
% \centering
% \includegraphics[width=0.9\linewidth]{fig/Implementation/pct2.pdf}
% \caption{\textcolor{red}{Offset-Attention (from \cite{guo2021pct}).}}
% \label{fig:cls-offset}
% \end{figure*}

\subsection{High-level Task}
\label{subsec:4.1}
In the field of 3D point cloud processing, high-level tasks usually include: classification \& segmentation~\cite{xie2018attentional,yang2019modeling,yan2020pointasnl,zhao2021point,gao2022lft,qiu2022geometric,wu2021centroid,han2021dual,han2021point,guo2021pct,yu20213d,zhang2021pvt,fu2022distillation,cheng2021patchformer,yu2021pointbert,lai2022stratified, wang2022local, liu2022group, yang2022mil, park2022fast, zhang2021u}, object detection~\cite{xie2020mlcvnet,pan20213d,liu2021group,misra2021end,sheng2021improving,mao2021voxel,chen2022pq,he2022voxset,zhang2022cat,wangbridged, yuan2021temporal}, tracking~\cite{cui20213d,zhou2021pttr,jiayao2022real}, registration~\cite{wang2019deep,wang2022storm,fischer2021stickypillars,fu2021robust,chen2021full,trappolini2021shape,qin2022geometric,yew2022regtr} and so on.
Here, we started by introducing classification \& segmentation tasks, which are very common and fundamental research topics in the field of 3D computer vision.

\subsubsection{Classification \& Segmentation}

Similar to image classification~\cite{krizhevsky2012imagenet,simonyan2014very,he2016deep,huang2017densely}, 3D point cloud classification methods aim at classifying the given 3D shapes into specific categories, such as chair, bed and sofa for indoor scenes, and pedestrian, cyclist and car for outdoor scenes.
In the field of 3D point cloud processing, since the encoders of segmentation networks are usually developed from classification networks, we introduce these tasks together.

Xie et al.~\cite{xie2018attentional}, for the first time, introduced the self-attention mechanism into the task of point cloud recognition. 
Inspired by the success of shape context~\cite{belongie2002shape} in shape matching and object recognition, the authors first transformed the input point cloud into a form of shape context representation. This representation was comprised of a set of concentric shell bins.
Based on the proposed novel representation, they then introduced the ShapeContextNet (SCN) to perform point feature extraction.
To automatically capture the rich local and global information, a dot-product self-attention module was further applied to the shape context representation, resulting in the Attentional ShapeContextNet (A-SCN).

% Considering the disability of PointNet~\cite{qi2017pointnet} in capturing local information, Chen et al.~\cite{CHEN2021122} proposed to apply the graph attention network~\cite{velivckovic2017graph} to extract local geometric features, resulting in their GAPNet.
% To this end, they designed a GAPLayer (multi-head graph attention-based point network layer) to replace the MLP layer in PointNet.
% The basic idea is to pay different attention to different neighboring points by regarding each point as a node and taking the connection between neighboring points as edges in graphs.
% Specifically, a self-attention operation and a neighboring-attention operation are proposed to encode self-geometric information and local (neighboring)-geometric information respectively.
% As a early attempt that leverages 

% Instead of employing the attention mechanism into enhancing feature representation of local neighboring points (i.e., local context), the following methods 

%%%%%% Figure 7
% \begin{figure}[t]
% \centering
% \includegraphics[width=0.95\linewidth]{fig/classification/pt2.pdf}
% \caption{Flow chart of the Point Transformer layer (from~\cite{zhao2021point}).}
% \label{fig:cls-point-transformer}
% \end{figure}

% As illustrated in Fig.~\ref{fig:cls-point-transformer}, 
Inspired by the self-attention networks in image analysis~\cite{ramachandran2019stand,zhao2020exploring} and NLP~\cite{devlin2018bert}, Zhao et al.~\cite{zhao2021point} designed a vector attention-based Point Transformer layer.
A Point Transformer block was constructed on the basis of the Point Transformer layer in a residual fashion.
% As shown in Fig. \ref{fig:pointtransformer}, 
The encoder of PT was constructed with only Point Transformer blocks, pointwise transformations and pooling operation for point cloud classification. Moreover, PT also used a U-Net structure for point cloud segmentation, where the decoder was designed to be symmetrical with the encoder. It presented a Transition Up module to recover the original point cloud with semantic features from the downsampled point set. Such module consisted of a linear layer, batch normalization, ReLU, and trilinear interpolation for feature mapping. Additionally, a skip connection between the encoder block and the corresponding decoder block was introduced to facilitate backpropagation.
With these carefully designed modules, PT became the first model that reached over $70\%$ mIoU ($70.4\%$) for semantic segmentation on Area 5 of the S3DIS dataset~\cite{armeni20163d}.
As for the task of shape classification on the ModelNet40 dataset, Point Transformer also achieved $93.7\%$ overall accuracy.

As illustrated in Sec. \ref{subsec:2.1}, Point-BERT~\cite{yu2021pointbert} was proposed to pre-train pure Transformer-based models with a Mask Point Modeling (MPM) task for point cloud classification. It was inspired by the concept of BERT~\cite{devlin2018bert} and masked autoencoder~\cite{he2021masked}.
Specifically, a point cloud was first divided into several local point patches. Then a mini-PointNet was utilized to get the embedded feature (which can be regarded as tokens) for each patch.
Like~\cite{yu2021pointbert}, some tokens were randomly discarded (masked) and the rest were fed to the Transformer network, to recover the masked point tokens.
This training procedure was entirely self-supervised.
With 8192 points as input, Point-BERT achieved $93.8\%$ overall accuracy on ModelNet40~\cite{wu20153d}.

Zhang et al.~\cite{zhang2021pvt} proposed a pure Transformer-based point cloud learning backbone, taking 3D voxels as the input, termed Point-Voxel Transformer (PVT). 
% The voxel form is an efficient representation of point cloud data~\cite{qi2016volumetric,zhou2018voxelnet,wang2019voxsegnet}. 
Inspired by the recent Swin Transformer~\cite{liu2021swin}, a Sparse Window Attention (SWA) operation was designed to perform the self-attention within non-overlapping 3D voxel windows in a shifting-window configuration.
A relative-attention (RA) operation was also introduced to compute fine-grained features of points.
With the two aforementioned modules, PVT could take advantage of both point-based and voxel-based structures with one pure Transformer architecture.
Similarly, Lai et al.~\cite{lai2022stratified} proposed Stratified Transformer to explicitly encode global contexts. 
It also extended Swin Transformer~\cite{liu2021swin} to point cloud processing by 3D voxelization. 
The main difference from PVT is that Stratified Transformer took both dense local points and sparse distant points as the $key$ vectors for each $query$ vector. 
This operation was beneficial to message passing among cubic windows and as well as to global information capturing. 
Both PVT and Stratified Transformer achieved $86.6\%$ pIoU for part segmentation on ShapeNet dataset.
However, Stratified Transformer performed better for semantic segmentation, surpassing PVT by $4.7\%$ mIoU on the S3DIS dataset.

\subsubsection{Object Detection}
Thanks to the popularization of 3D point cloud scanners, 3D object detection is becoming a more and more popular research topic.
Similar to the 2D object detection task, 3D object detectors aim to output 3D bounding boxes with point clouds as input data.
Recently, Carion et al.~\cite{carion2020end} introduced the first Transformer-based 2D object detector, DETR. It proposed to combine Transformers and CNNs to eliminate non-maximum suppression (NMS). 
Since then, Transformer-related works have also shown a flourishing growth in the field of point cloud-based 3D object detection.

% \begin{figure*}[t]
% \centering
% \includegraphics[width=\linewidth]{fig/3DETR.pdf}
% \caption{3DETR for 3D object detection from point clouds (from \cite{misra2021end}).}
% \label{fig:3detr}
% \end{figure*}

On the basis of VoteNet~\cite{qi2019deep}, Xie et al.~\cite{xie2020mlcvnet,xie2021vote}, for the first time, introduced the self-attention mechanism of Transformers into the task of 3D object detection in indoor scenes. They proposed the Multi-Level Context VoteNet (MLCVNet) to improve detection performance by encoding contextual information.
In their papers, each point patch and vote cluster were regarded as tokens in Transformers. 
Then the self-attention mechanism was utilized to strengthen the corresponding feature representations via capturing relations within point patches and vote clusters, respectively.
Due to the integration of the self-attention modules, MLCVNet achieved better detection results than its baseline model on both ScanNet~\cite{dai2017scannet} and SUN RGB-D datasets~\cite{song2015sun}.
PQ-Transformer~\cite{chen2022pq} was proposed to detect 3D objects and predict room layouts simultaneously.
which was also based on VoteNet. It utilized a Transformer decoder to enhance proposal features.
With the assistance of room layout estimation and refined features by the Transformer decoder, PQ-Transformer attained a mAP@0.25 of $67.2\%$ on ScanNet.

% \textcolor{red}{To achieve effective feature learning, Pan et al.~\cite{pan20213d} proposed a pure Transformer-based backbone, Pointformer, whose architecture followed the U-Net fashion.
% As explained in Sec. \ref{subsec:2.1}, three types of Transformer-based blocks were introduced in the Pointformer paper: LT, LGT and GT.
% Similar to MLCVNet, these blocks were designed to enhance feature representative with the aid of encoding long-range dependencies of Transformers.
% The proposed Pointformer improved detection performance on both indoor datasets (SUN RGB-D~\cite{song2015sun} and ScanNet V2~\cite{dai2017scannet}) and outdoor datasets (nuScenes~\cite{caesar2020nuscenes} and KITTI~\cite{geiger2012we}).}
% when it is used as a backbone for existing detection models.

Aforementioned methods employed the hand-crafted grouping scheme, obtaining features for object candidates by learning from points within the corresponding local regions. However, Liu et al.~\cite{liu2021group} argued that the point grouping operation within limited regions tended to hinder the performance of 3D object detection.
Thus, they presented a group-free framework with the aid of the attention mechanism in Transformers.
The core idea was that the features of an object candidate should come from all the points in the given scene, instead of a subset of the point cloud.
After obtaining object candidates, their method first leveraged a self-attention module to capture contextual information between the object candidates.
They then designed a cross-attention module to refine the object features with the information of all the points.
With the improved attention stacking scheme, their detector achieved the mAP@0.25 of $69.1\%$  on the ScanNet dataset.

Inspired by DETR~\cite{carion2020end} in 2D object detection, an end-to-end 3D DEtection Transformer network, termed 3DETR~\cite{misra2021end}, was first proposed to formulate 3D object detection as a set-to-set problem.
Borrowing ideas from both DETR~\cite{carion2020end} and VoteNet~\cite{qi2019deep}, 3DETR was designed in the general encoder-decoder fashion.
% as illustrated in Fig.~\ref{fig:3detr}.
In the encoder part, sampled points and the corresponding features extracted by MLP were directly fed into a Transformer block for feature refinement.
In the decoder part, these features went through a parallel Transformer-fashion decoder and were turned into a set of object candidate features.
These object candidate features were finally used to predict 3D bounding boxes.
3DETR improved on VoteNet by $9.5\%$ $AP_{50}$ and $4.6\%$ $AP_{25}$ on ScanNetV2 and SUN RGB-D respectively.

% \textcolor{red}{Images can provide complementary information for object detection from 3D point clouds~\cite{qi2020imvotenet}.
% Bridged Transformer (BrT)~\cite{wangbridged} focused on exploring the multi-modal fusion strategy. It fused point clouds and images for 3D object detection in indoor scenes.
% Considering the differences in structure between point clouds and images, BrT did not directly fuse them by simply applying the attention mechanism.
% Instead, point and image patch tokens were both fed into the Bridged Transformer layers. Then object queries were utilized to bridge information communication between points and images.
% Benefiting from this bridged design, BrT reached $71.3\%$ mAP@0.25 on ScanNetV2 validation set.}

% \begin{figure*}[t]
% \centering
% \includegraphics[width=\linewidth]{fig/PTTR.pdf}
% \caption{Flow chart of PTTR (from \cite{zhou2021pttr}).}
% \label{fig:pttr}
% \end{figure*}

Apart from the aforementioned methods focusing on indoor scenes, Sheng et al.~\cite{sheng2021improving} proposed a Channel-wise Transformer based two-stage framework (CT3D) to improve 3D object detection performance in outdoor LiDAR point clouds.
The input of the channel-wise Transformer came from a Region Proposal Network (RPN). Moreover, the Transformer network consisted of two sub-modules: the proposal-to-point encoding module and the channel-wise decoding module.
The encoding module first took the proposals and their corresponding 3D points as input. Then it extracted the refined point features through a self-attention-based block.
The channel-wise decoding module transformed the extracted features from the encoder module into a global representation through a channel-wise re-weighting scheme. Finally, Feed-Forward Networks (FFNs) were performed for detection predictions.
As such, CT3D achieved $81.77\%$ AP in the moderate car category on the KITTI test set.

% Different from those methods working directly on point clouds, Mao et al.~\cite{mao2021voxel} first transformed the point clouds into the voxel-based representation, and then proposed a Voxel Transformer (VoTr) backbone for 3D object detection.
% They presented the submanifold voxel module and the sparse voxel module to extract features from non-empty and empty voxels respectively.
% In both two modules, the Local Attention and Dilated Attention operation are designed, on the basis of multi-head attention, to maintain low computational consumption for numerous voxels.
% As stated, the proposed VoTr can be integrated into most 3D detectors whose data representative format is voxel-based.
% To tackle the computation issue of Transformers in voxel-based outdoor 3D detectors, He et al.~\cite{he2022voxset} proposed Voxel Set Transformer (VoxSeT) to detect outdoor objects in a set-to-set fashion.
% Based on the low-rank characteristic of the self-attention matrix, a voxel-based attention (VSA) module is designed by assigning a set of trainable ``latent codes" to each voxel, which is inspired by the induced set attention blocks in Set Transformer~\cite{lee2019set}.

% Based on the observation that the association between LiDAR points and images pixels established by calibration matrices is insufficient, 
In a similar paradigm to DETR~\cite{carion2020end}, a LiDAR and Camera fusion based 3D object detector based on Transformers was proposed in \cite{bai2021pointdsc}, called TransFusion.
In TransFusion, the attention mechanism was employed to adaptively fuse features from images. It aimed to relieve the problem of bad association between LiDAR points and image points established by calibration matrices.
CAT-Det~\cite{zhang2022cat} was also proposed to fuse LiDAR point clouds and RGB images more efficiently for 3D object detection performance boosting.
A Pointformer and an Imageformer were first introduced in the branches of the point cloud and image respectively to extract multi-modal features.
A Cross-Modal Transformer (CMT) module was then designed to combine the features from the aforementioned two streams.
With the performance of $67.05\%$ mAP on the KITTI test split, CAT-Det became the first multi-modal solution that significantly surpassed LiDAR-only ones.

% \textcolor{red}{Temporal-Channel TRansformer (TCTR)~\cite{yuan2021temporal} was proposed to process 3D Lidar-based video for effective object detection in autonomous driving. The key idea was based on the observation that adjacent frames can provide contextual information to the current frame.
% Instead of merely taking the current frame $t$ point cloud as input, it proposed to include the former $T$ frames to assist in object detection for frame $t$.
% Specifically, the input raw point clouds were first transformed into images. Then TCTR was designed to extract and aggregate features from multiple frames, by encoding the temporal-channel domain and spatial-wise relationships along with the continuous frames.}

\subsubsection{Object Tracking}
3D object tracking takes two point clouds (i.e., a template point cloud and a search point cloud) as input. It outputs 3D bounding boxes of the target (template) in the search point cloud.
It involves feature extraction of point clouds and feature fusion between template and search point clouds.

Cui et al.~\cite{cui20213d} argued that most existing tracking approaches did not consider the attention changes of object regions during tracking.
According to them, different regions in the search point cloud should contribute different importance to the feature fusion process.
Based on this observation, they presented a LiDAR-based 3D Object Tracking with a TRansformer network (LTTR). This method was able to improve the feature fusion of template and search point clouds by capturing attention changes over tracking time.
Specifically, they first built a Transformer encoder to improve the feature representation of template and search point clouds separately.
% , by encoding relations within regions of the point clouds. 
Then the cross-attention mechanism was employed to build a Transformer decoder. It could fuse features from the template and search point clouds by capturing relations between the two point clouds.
Benefiting from the Transformer-based feature fusion between the template and search point clouds, LTTR reached $65.8\%$ mea Precision on KITTI tracking dataset.
Zhou et al.~\cite{zhou2021pttr} also proposed a Point Relation Transformer (PRT) module to improve feature fusion in their coarse-to-fine Point Tracking TRansformer (PTTR) framework.
% , as illustrated in Fig.~\ref{fig:pttr}.
Similar to LTTR, PRT employed self-attention and cross-attention to encode relations within and between point clouds respectively.
The difference is that PRT utilized the Offset-Attention~\cite{guo2021pct} to relieve the impact of noise data. 
Finally, PTTR surpassed LTTR by $8.4\%$ and $10.4\%$ in terms of average Success and Precision, and became a new SOTA on the KITTI tracking benchmark.

Unlike the two aforementioned approaches which focused on the feature fusion step, Shan et al.~\cite{jiayao2022real} introduced a Point-Track-Transformer (PTT) module to enhance the feature representation after the feature fusion step.
Features from the fusion step and the corresponding point coordinates were both mapped into the embedding space.
A position encoding block was also designed to capture positional features using the KNN algorithm and an MLP layer.
With the aforementioned two embedded semantic and positional features as input, a self-attention block was finally applied to obtain more representative features.
To verify the effectiveness of the proposed PTT, the authors integrated it into the seeds voting and proposal generation stages of the P2B~\cite{qi2020p2b} model resulting in the PTT-Net.
PTT-Net improved P2B by $9.0\%$ in terms of Precision on KITTI for the car category.

% \subsubsection{Reconstruction.}

% xxx...

% \begin{figure*}[t]
% \centering
% \includegraphics[width=\linewidth]{fig/REGTR-registration.pdf}
% \caption{REGTR: End-to-end Point Cloud Correspondences with Transformers (from \cite{yew2022regtr}).}
% \label{fig:regtr}
% \end{figure*}

\subsubsection{Registration}
Given two point clouds as input, the aim of point cloud registration is to find a transformation matrix to align them.

Deep Closest Point (DCP) model proposed in~\cite{wang2019deep} introduced the Transformer encoder into the task of point cloud registration.
As usual, the input unaligned point clouds were first sent to a feature embedding module, such as PointNet~\cite{qi2017pointnet} and DGCNN~\cite{wang2019dynamic}, to transfer 3D coordinates into a feature space.
A standard Transformer encoder was then applied to perform context aggregation between two embedded features.
Finally, DCP utilized a differentiable Singular Value Decomposition (SVD) layer to compute the rigid transformation matrix.
DCP was the first work that employed the Transformer model to improve the feature extraction of point clouds in registration.
With the same paradigm, STORM~\cite{wang2022storm} also deployed Transformer layers to refine the point-wise features extracted by EdgeConv~\cite{wang2019dynamic} layers, capturing the long-term relationship between point clouds.
It achieved better performance than DCP for partial registration on ModelNet40 dataset.
Similarly, Fischer et al.~\cite{fischer2021stickypillars} leveraged multi-head self- and cross-attention mechanisms to learn contextual information between target and source point clouds. Their method focused on processing outdoor scenes, e.g., the KITTI dataset~\cite{geiger2012we}.

To find more robust correspondences between two point clouds, Fu et al.~\cite{fu2021robust} presented the first deep graph matching-based framework (RGM) to perform robust point cloud registration, which was less sensitive to outliers.
During the graph establishment, they employed Transformer encoders to obtain the soft edges of two nodes within a graph.
With the generated soft graph edges, better correspondences could be obtained for the overlapping parts when registering partial-to-partial point clouds.
The effectiveness of the proposed Transformer-based edge generator was demonstrated by the ablation stud where the performance dropped on ModelNet40 when replacing the edge generator with either full connection edges or sparse connection edges.

% \textcolor{red}{To address the problem of indistinct feature extraction caused by the shallow-wide Transformer architecture,
% % Considering that previous Transformer-based point cloud registration methods suffer from indistinct feature extraction due to the shallow-wide architecture of Transformer,
% Deep Interaction Transformer (DIT)~\cite{chen2021full} was proposed.
% It contained three novel carefully designed modules to perform feature extraction and correspondence confidence evaluation.
% To obtain good representations of each input point cloud, a Point Cloud Structure Extractor (PSE) was introduced. It employed the Transformer encoder to model global relations, and proposed a Local Feature Integrator (LFI) to encode structural information.
% The extracted features $(F_X, F_Y)$ of two input point clouds were then fed into a deep-narrow Point Feature Transformer (PFT), to establish associations. 
% Moreover, a positional encoding network was inserted to encode relative position information between points.
% As such, feature representations $(\Psi_X, \Psi_Y)$ with richer information can be obtained.
% Given two features and the established correspondences, a Geometric Matching-based Correspondence Confidence Evaluation (GMCCE) was designed to filter out bad correspondence with low confidence values.
% With more representative features extracted by the full Transformer network, DIT outperformed previous methods, achieving $1.1e-8$ in terms of $t_{MAE}$ on clean point clouds of ModelNet40~\cite{wu20153d}.}

Recently, Yew et al.~\cite{yew2022regtr} argued that explicit feature matching and outlier filtering via RANSAC in point cloud registration can be replaced with attention mechanisms.
% As illustrated in Fig.~\ref{fig:regtr}, 
They designed an end-to-end Transformer framework, termed REGTR, to directly find point cloud correspondences.
In REGTR, point features from a KPconv~\cite{thomas2019kpconv} backbone were fed into several multi-head self- and cross-attention layers for comparing source and target point clouds.
% to capture relationships within and between source and target point clouds.
With the aforementioned simple design, REGTR became the current state-of-the-art point cloud registration method on the ModelNet40~\cite{wu20153d} and 3DMatch~\cite{zeng20173dmatch} datasets.
Similarly, GeoTransformer~\cite{qin2022geometric} also utilized self- and cross-attention to find robust superpoint correspondences.
In terms of Registration Recall, both REGTR and GeoTransformer achieved $92.0\%$ on the 3DMatch dataset.
However, GeoTransformer surpassed REGTR by $10.2\%$ on the 3DLoMatch~\cite{huang2021predator} dataset.

\subsubsection{Point Cloud Video Understanding}
The 3D world around us is dynamic and consistent in time, which cannot be fully represented by traditional single-frame and fixed point clouds.
In contrast, point cloud videos, a set of point clouds captured in a fixed frame rate, could be a promising data representation of dynamic scenes in the real world.
Understanding dynamic scenes and dynamic objects is important for the application of point cloud models to many real-world scenarios.
Point cloud video understanding involves processing a time sequence of 3D point clouds.
Thus, the Transformer architecture could be a promising choice to process point cloud videos, since they are good at dealing with global long-range interactions.

% \begin{figure*}[t]
% \centering
% \includegraphics[width=\linewidth]{fig/PU-Transformer.pdf}
% \caption{Flow chart of PU-Transformer (from \cite{qiu2021pu}).}
% \label{fig:pu-transformer}
% \end{figure*}

Based on such observation, P4Transformer~\cite{fan2021point} was proposed to process point cloud videos for action recognition.
To extract the local spatial-temporal  features of a point cloud video, the input data were first represented by a set of spatial-temporal local areas. Then a point 4D convolution was used to encode features for each local area.
After that, the P4Transformer authors introduced a Transformer encoder to receive and integrate the features of local areas via capturing long-range relationships across the entire video.
P4Transformer has been successfully applied to the task of 3D action recognition and 4D semantic segmentation from point clouds. 
It achieved higher results than PointNet++-based methods on many benchmarks (e.g., the MSR-Action3D~\cite{li2010action}, the NTU RGB+D 60~\cite{shahroudy2016ntu} and 120~\cite{liu2019ntu} datasets for 3D action recognition, and the Synthia 4D~\cite{choy20194d} dataset for 4D semantic segmentation).
It demonstrated the effectiveness of Transformers on point cloud video understanding.

\subsection{Low-level Task}

The input data of low-level tasks is usually the raw scanned point cloud with occlusion, noise, and uneven densities.
Thus, the ultimate goal of low-level tasks is to get a high-quality point cloud, which could benefit high-level tasks.
Some typical low-level tasks include point cloud downsampling~\cite{wang2022lightn}, upsampling~\cite{qiu2021pu}, denoising~\cite{gao2022reflective,xu2022tdnet}, completion~\cite{yu2021pointr,xiang2021snowflakenet,chen2021transsc,lin2021pctma,yan2022shapeformer,liu2022point, huang20223dpctn, wen2022pmp}.

% \subsubsection{Sampling.}

\subsubsection{Downsampling}
Given a point cloud with $N$ points, downsampling methods aim at outputting a smaller size of point cloud with $M$ points, while retaining the geometric information of the input point cloud.
Leveraging the powerful learning ability of Transformers, LighTN~\cite{wang2022lightn} was proposed to downsample point clouds in a task-oriented manner. 
As mentioned in Sec.~\ref{subsec:2.3}, it first removed the position encoding, then used a small-size shared linear layer as the embedding layer.
Moreover, the MSA module was replaced with a single head self-correlation layer.
Experimental results demonstrated the aforementioned strategies significantly reduced the computational cost.
$86.18\%$ classification accuracy could still be attained while only 32 points were sampled.
Moreover, the lightweight Transformer network was designed as a detachable module, which can be easily inserted into other neural networks.

\subsubsection{Upampling}
Contrary to downsampling, upsampling methods aim to restore missing fine-scale geometric information by outputting a point cloud of bigger size than the input point cloud \cite{Wei-AGConv2022}. The upsampled points are expected to reflect realistic geometry and lie on the surfaces of the objects represented by the given sparse point clouds.
PU-Transformer~\cite{qiu2021pu} was the first work to apply the Transformer-based model to point cloud upsampling.
% , as illustrated in Fig.~\ref{fig:pu-transformer}.
The authors designed two novel blocks for the PU-Transformer.
The first block was the Positional Fusion block (PosFus), which aimed at capturing local position-related information.
The second one was the Shifted Channel Multi-head Self-Attention (SC-MSA) block. It was designed to address the lack of connection between the outputs of different heads in conventional MSA. See the SC-MSA in Sec. \ref{sec:5} for more details.
PU-Transformer showed the promising potential of Transformer-based models in point cloud upsampling.

\subsubsection{Denoising}

Denoising takes point clouds corrupted by noise as input, and outputs clean point clouds by utilizing the local geometry information.
TDNet~\cite{xu2022tdnet} was first proposed for point cloud denoising. Taking each point as a word token, it improved the NLP Transformer~\cite{vaswani2017attention} making it suitable for point cloud feature extraction.
The Transformer-based encoder mapped the input point cloud into a high-dimensional feature space and learned the semantic relationship among points.
With the extracted feature from the encoder, the latent manifold of the noisy input point cloud can be obtained.
Finally, a clean point cloud can be generated by sampling each patch manifold.

Another category of point cloud denoising method is to filter out noise points directly from the input point clouds.
For instance, some Lidar point clouds could contain a huge number of virtual (noise) points. These points are produced by the specular reflections of glass or other kinds of reflective materials.
To detect these reflective noise points, Gao et al.~\cite{gao2022reflective} first projected the input 3D LiDAR point cloud into a 2D range image. Then a Transformer-based auto-encoder network was employed to predict a noise mask to indicate the points coming from reflection.

% \begin{figure*}[t]
% \centering
% \includegraphics[width=\linewidth]{fig/PoinTr.pdf}
% \caption{Diverse Point Cloud Completion with Geometry-Aware Transformers (from \cite{yu2021pointr}).}
% \label{fig:pointr}
% \end{figure*}

\subsubsection{Completion}
In most 3D practical applications,
% using 3D point cloud scanners,
it is usually difficult to obtain complete point clouds of objects or scenes due to occlusion from other objects or self-occlusion. This issue makes point cloud completion an important low-level task in the field of 3D vision. 
%The complete point cloud contains more geometrical information about objects, which can be used to help computers understand the physical world better.

%PointTr(2021-04)
% As illustrated in Fig.~\ref{fig:pointr}, 
PoinTr proposed in~\cite{yu2021pointr}, for the first time, converted point cloud completion to a set-to-set translation task.
Specifically, the authors claimed that the input point cloud can be represented by a set of groups of local points, termed ``point proxies".
Taking a sequence of point proxies as input, a geometry-aware Transformer block was carefully designed to generate the point proxies of the missing parts.
In a coarse-to-fine fashion, FoldingNet~\cite{yang2017foldingnet} was finally employed to produce points based on the predicted point proxies.
The geometry-aware Transformer block was a self-contained module, which can capture both the semantic and geometric relationship among points.
PoinTr attained $8.38$ Average $L_{1}$ Chamfer Distance (CD) on the PCN dataset~\cite{yuan2018pcn}.

In contrast with PointTr, Xiang et al.~\cite{xiang2021snowflakenet} proposed to formulate the task of point cloud completion as the growth of 3D points in a snowflake-like fashion.
Based on this insight, SnowflakeNet was presented to focus on recovering fine geometric details, such as corners, sharp edges and smooth regions, of the complete point cloud.
The core idea was to combine Snowflake Point Deconvolution (SPD) layer with the skip-Transformer to better guide the point splitting process.
SPD could generate multiple points from any single one.
Skip-Transformer was capable of capturing both contexts and spatial information from the given point and the generated points.
With the skip-Transformer integrated, the SPD layers were capable of modeling structure characteristics, thus producing more compact and structured point clouds.
Benefiting from SPD and the skip-Transformer, SnowflakeNet surpassed PoinTr by $1.17$ Average $L_{1}$ Chamfer Distance (CD) on the PCN dataset \cite{yuan2018pcn}.

% \textcolor{red}{Due to the partial scanned data, robotic grasping methods often suffer from wrong grasping estimation.
% To solve this issue, a robotic grasping-oriented shape completion model was presented, termed TransSC~\cite{chen2021transsc}.
% In TransSC, a Transformer-based Multi-Layer Perception (TMLP) module was designed to extract better point-wise feature representations. Then a manifold-based decoder was employed to produce the complete point clouds by decoding the point features. PCTMA-Net~\cite{lin2021pctma} also leveraged a Transformer encoder to improve feature representation. Similar to TransSC, the authors claimed it claimed that the Transformer-based embedding network can extract more discriminate features for each point than MLP-based networks.
% Liu et al.~\cite{liu2022point} also integrated self-attention and cross-attention to enhance feature extraction in their proposed dynamic Transformer-based point cloud completion framework.}

Instead of working directly on the point cloud, ShapeFormer proposed in~\cite{yan2022shapeformer} introduced a novel 3D sparse representation named the Vector Quantized Deep Implicit Functions (VQDIF). It converted the 3D point cloud to a set of discrete 2-tuples consisting of the coordinate and the quantized feature index. On this basis, a VQDIF encoder and decoder were designed to perform transformation between the 3D point cloud and the proposed 2-tuples.
The sequences of 2-tuples features from partial observations were fed into a Transformer-based autoregressive model to generate complete feature sequences.
Then these sequences were projected to a feature grid via the VQDIF decoder.
Finally, a 3D-Unet \cite{cciccek20163d} was employed to generate local deep implicit functions of objects' whole shapes.

%% file: 5_Variants.tex
\section{3D Self-attention Variants}
\label{sec:5}

\begin{figure*}[htbp]
\centering
\includegraphics[width=0.8\linewidth]{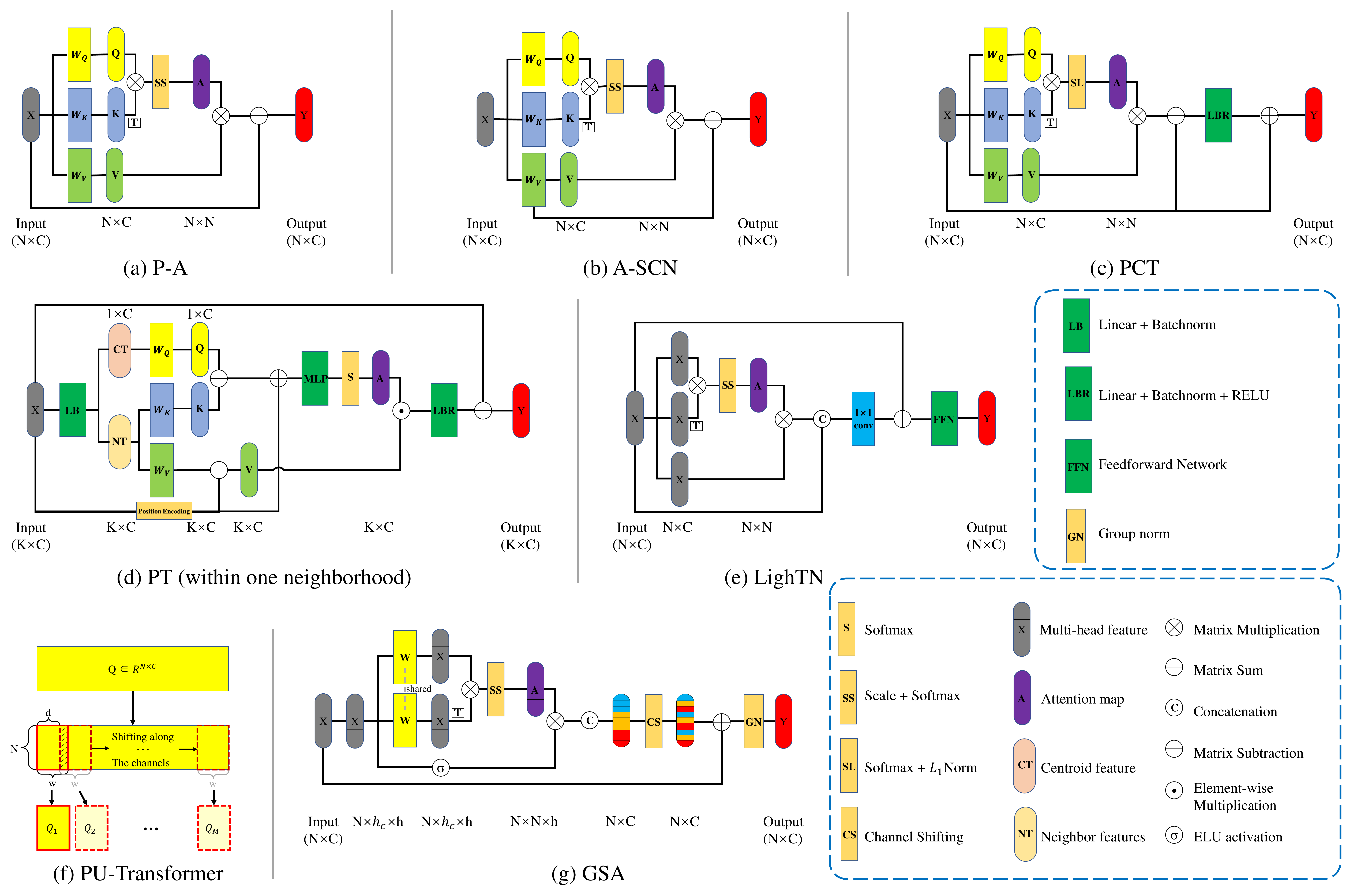}
\caption{Architectures of Point-wise Variants.}
\label{fig:Point-wise Variants}
\end{figure*}

\begin{figure*}[htbp]
\centering
\includegraphics[width=0.75\linewidth]{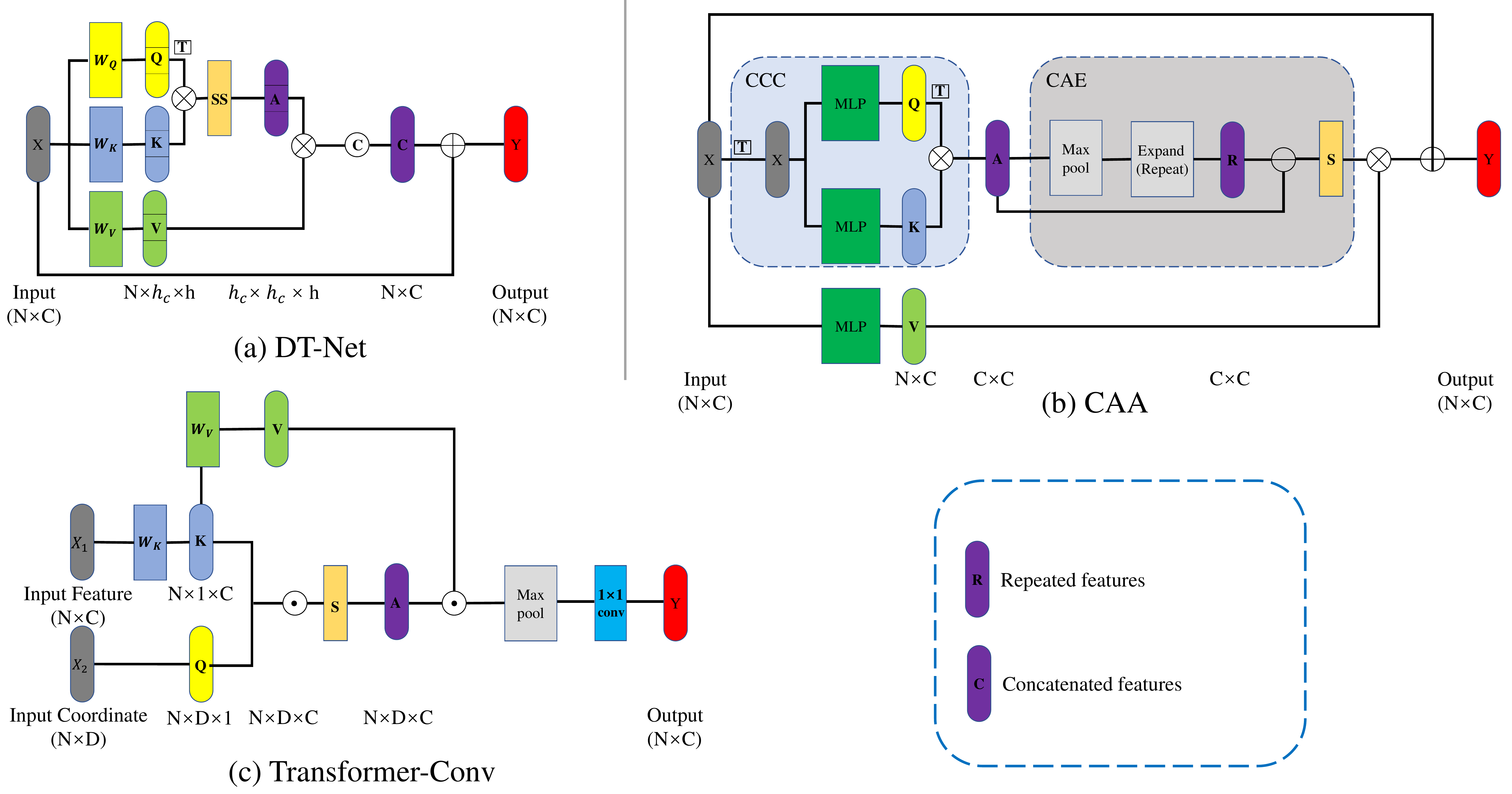}
\caption{Architectures of Channel-wise Variants.}
\label{fig:Channel-wise Variants}
\end{figure*}

Based on the standard self-attention module, there are many variants which were designed to improve the performance of Transformers in 3D point cloud processing, as shown in Figs. \ref{fig:Point-wise Variants} and \ref{fig:Channel-wise Variants}. As stated in Sec. \ref{subsec:2.2}, we categorize the relevant variants into two classes: \textit{Point-wise Variants} and \textit{Channel-wise Variants}.

\subsection{Point-wise Variants}
\label{subsec:point-wise variant}
P-A \cite{feng2020point} (Fig. \ref{fig:Point-wise Variants}(a)) and  A-SCN \cite{xie2018attentional} (Fig. \ref{fig:Point-wise Variants}(b)) have different residual structures in their Transformer encoders. The former strengthened the connection between the output and input of the module, while the later established the relationship between the output and the $Value$ matrix of the module. Relevant experiments have demonstrated that the residual connection facilitated model convergence \cite{xie2018attentional}. 

Inspired by the Laplacian matrix $L = D - E$ in Graph convolution networks \cite{bruna2013spectral}, the PCT paper \cite{guo2021pct} further proposed an Offset-Attention module (Fig. \ref{fig:Point-wise Variants}(c)). This module calculated the offset (difference) between the Self-Attention (SA) features and the input features $X$ by matrix subtraction, which was analogous to a discrete Laplacian operation. Additionally, it refined the normalization of the similarity matrix by replacing \textit{Scale} + \textit{Softmax} (SS) with \textit{Softmax}  + \textit{$L_{1}$ Norm} (SL) operation. It was able to sharpen the attention weights and reduce the influence of noise. Based on the Offset-Attention, Zhou et al. \cite{zhou2021pttr} proposed a Relation Attention Module (RAM) which had a similar structure as the Offset-Attention module. The difference was that it first projected $Query$, $Key$ and $Value$ matrices into latent feature spaces by linear layers. Then, instead of generating the $Attentionmap$ by multiplying the $Query$ and $Key$ matrices directly, it applied the $L_{2}$ normalization to the $Query$ and $Key$ matrices. This operation prevented the few feature channels with extremely large magnitudes from overpowering the rest. Ablation experiments in \cite{zhou2021pttr} demonstrated that the $L_{2}$ normalization was able to improve the model performance.

PT \cite{zhao2021point} (Fig. \ref{fig:Point-wise Variants}(d)) introduced the vector subtraction attention operator to its Transformer network, replacing the commonly-used scalar dot-product attention. Compared with the scalar attention, vector attention is more expressive since it supports adaptive modulation of individual feature channels, as opposed to whole feature vectors. This kind of expression appears to be very beneficial in 3D data processing \cite{zhao2021point}. Point Transformer utilized the subtraction-form vector attention to achieve the local feature aggregation. The attention map was generated by simply building the connections between the centroid feature and its neighbor feature, instead of measuring the similarity between any two point features within a neighborhood. 
Additionally, the 3D Convolution-Transformer Network (3DCTN) paper \cite{lu20223dctn} conducted a detailed investigation on self-attention operators in 3D Transformers, including the scalar attention and different forms of vector attention.

As mentioned in Sec. \ref{subsec:2.3}, LighTN \cite{wang2022lightn} presented a self-correlation module, to reduce the computational cost. As shown in Fig. \ref{fig:Point-wise Variants}(e), it eliminated the projection matrices, $W_{Q}$, $W_{K}$, and $W_{V}$ simultaneously in the self-attention mechanism. Only the input self-correlation parameters were used to generate attention features. According to Eq. \ref{self-correlation}, the self-correlation mechanism generates a symmetry attention map $X \cdot X^{T}$, which satisfies the permutation invariance in point cloud processing \cite{wang2022lightn}. The authors also conducted a series of ablation studies, removing different projection matrices, to demonstrate the effectiveness of the proposed self-correlation mechanism.

PU-Transformer \cite{qiu2021pu} proposed the SC-MSA block to improve the MSA mechanism. Specifically, despite the rich information captured by MSA, only feature dependencies within the same head can be estimated. I.e., MSA lacked information propagation between different heads. To address this issue, as shown in Fig. \ref{fig:Point-wise Variants}(f), PU-Transformer applied a window (dashed square) shift along the channels to ensure that any two consecutive splits had a uniform overlap area. Compared with the independent splits of the standard MSA, SC-MCA is able to enhance the channel-wise relations in the output features.

GSA proposed in \cite{yang2019modeling} had two improvements compared to the standard MSA. The first one was that GSA was a parameter-efficient self-attention mechanism. It used a shared projection matrix $W$ to generate the $Query$ and $Key$ matrices, and used a non-linearity $\sigma$ to generate the $Value$ matrix:
\begin{equation}
\begin{aligned}
Attn_{\sigma}(X) =  softmax(\frac{Q K^{T}}{\sqrt{C}}) \sigma(X),
\end{aligned}
\end{equation}
where $Q = K = XW$ and $C$ is the dimension of $X$. With this, GSA is able to reduce the computational costs of the self-attention operation. 
The second one was that GSA introduced channel shuffle to MSA, which enhanced the information flow between heads. As shown in Fig. \ref{fig:Point-wise Variants}(g), unlike PU-Transformer \cite{qiu2021pu}, it re-grouped the channels by rewriting each point feature.

\subsection{Channel-wise Variants}

Dual Transformer Network (DT-Net) \cite{han2021dual} proposed the channel-wise MSA, applying the self-attention mechanism to the channel space. As shown in Fig. \ref{fig:Channel-wise Variants}(a), unlike the standard self-attention mechanism, the channel-wise MSA multiplied the transposed $Query$ matrix and $Key$ matrix. As such, the attention map could be generated to measure the similarities between different channels, as described in Eq. \ref{eq:2}.

As shown in Fig. \ref{fig:Channel-wise Variants}(b), the CAA module \cite{qiu2022geometric} utilized a similar approach to generate the similarity matrix between different channels. Moreover, it designed a CAE block to generate the affinity matrix, strengthening the connection between distinct channels and avoiding aggregating similar/redundant information. The $Value$ matrix was generated by an MLP layer, and the final feature map was obtained by multiplying the affinity matrix and $Value$ matrix. Additionally, the CAA module used a regular skip connection between the input and output feature map.

The Transformer-Conv module proposed in \cite{xu2021adaptive} learned the potential relationship between feature channels and coordinate channels. As shown in Fig. \ref{fig:Channel-wise Variants}(c), The $Query$ matrix and $Key$ matrix were generated by coordinates and features of the point cloud respectively. Then the similarity matrix was produced by a relation function $\beta$ (e.g., element-wise multiplication) and channel softmax operation. In contrast with the aforementioned methods, the $Value$ matrix in the Transformer-Conv module was generated from the $Key$ matrix by linear projection, followed by multiplying the similarity matrix and $Value$ matrix in an element-wise manner. Lastly, the final feature was generated by using a channel max-pooling and further $1 \times 1$ convolution.

%% file: 6_Comparison_and_Analysis.tex
\section{Comparison and Analysis}
\label{sec:6}

This section gives an overall comparison and analysis of 3D Transformers on several main-stream tasks, including classification, part segmentation, semantic segmentation and object detection.

%Comment: Table I,II,III titles too long%

%%%%%% classification on ModelNet40 %%%%%%%%%%%%%%%%%%%
\begin{table}[t!]
    \centering
    \caption{Comparative analysis between involved point cloud classification methods on the ModelNet40~\cite{wu20153d} dataset. OA means Overall Accuracy.
        All results quoted were taken from the cited papers.
        P = points, N = normals.}
    % \begin{tabular}{l|llrr}
    \begin{tabular}{l|lrr}
        \hline
        \textbf{Method}                               %& \textbf{Publication}  
        & \textbf{input} & \textbf{input size} & \textbf{OA(\%)} \\
        \hline
        \multicolumn{4}{c}{Non-Transformer} \\
        \hline
        PointNet~\cite{qi2017pointnet}             %& CVPR 2017
        & P       & $1024\times3$                & 89.2            \\
        PointNet++~\cite{qi2017pointnet++}          %& NIPS 2017  
        & P       & $1024\times3$                & 90.7            \\
        PointNet++~\cite{qi2017pointnet++}          %& NIPS 2017           
        & P, N    & $5120\times6$                & 91.9            \\
        PointWeb~\cite{zhao2019pointweb}            %& CVPR 2019        
        & P       & $1024\times3$                   & 92.3            \\
        SpiderCNN~\cite{xu2018spidercnn}            %& ECCV 2018   
        & P, N    & $1024\times6$                   & 92.4            \\
        PointCNN~\cite{li2018pointcnn}             % & NIPS 2018              
        & P       & $1024\times3$                   & 92.5            \\
        PointConv~\cite{wu2019pointconv}             % & CVPR 2019           
        & P, N       & $1024\times6$                   & 92.5            \\
        FPConv~\cite{lin2020fpconv}               %  & CVPR 2020            
        & P, N      & $1024\times6$                   & 92.5            \\
        Point2sequence~\cite{liu2019point2sequence} %& AAAI 2019             
        & P       & $1024\times3$                   & 92.6            \\
        DGCNN~\cite{wang2019dynamic}               % & TOG 2019          
        & P       & $1024\times3$                   & 92.9            \\
        KPConv~\cite{thomas2019kpconv}             % & ICCV 2019      
        & P       & $6800\times3$                   & 92.9            \\
        InterpCNN~\cite{mao2019interpolated}       % & ICCV 2019          
        & P       & $1024\times3$                   & 93.0            \\
        ShellNet~\cite{zhang2019shellnet}       % & ICCV 2019       
        & P       & $1024\times3$                  & 93.1  \\
        RSMix\cite{lee2021regularization}        %& CVPR 2021        
        & P       & $1024\times3$                   & 93.5   \\
        PAConv~\cite{xu2021paconv}       % & CVPR 2021      
        & P       & $1024\times3$                   & 93.9           \\
        RPNet~\cite{ran2021learning}        %& ICCV 2021  
        & P, N       & $1024\times6$                    & 94.1           \\
        CurveNet~\cite{xiang2021walk}       % & ICCV 2021 
        & P       & $1024\times3$                    & 94.2            \\
        PointMLP~\cite{ma2022rethinking}      %  & ICLR 2022
        & P       & $1024\times3$                   & 94.5            \\
        \hline
        \multicolumn{4}{c}{Attention/Transformer} \\
        \hline
        ShapeContextNet~\cite{xie2018attentional}   %& CVPR 2018            
        & P       & $1024\times3$                & 90.0            \\
        % 3D-ViT~\cite{fu2022distillation}            & Arxiv 2022              & P       & $1024\times3$                & 91.6            \\
        PATs~\cite{yang2019modeling}                % & CVPR 2019         
        & P       & $1024\times3$                & 91.7            \\
        % GAPNet~\cite{CHEN2021122}                   & Neurocomputing 2021     & P       & $1024\times3$                & 92.4            \\
        DT-Net~\cite{han2021dual}                  %  & Arxiv 2021          
        & P       & $1024\times3$                & 92.9            \\
        MLMSPT~\cite{han2021point}                 % & Arxiv 2021          
        & P       & $1024\times3$                & 92.9            \\
        % Cloud Transformers~\cite{mazur2021cloud}    & ICCV 2021               & P, N    & $1024\times6$                & 93.1            \\
        PointASNL~\cite{yan2020pointasnl}          % & CVPR 2020          
        & P, N    & $1024\times6$                & 93.2            \\
        PCT~\cite{guo2021pct}                     %  & CVMJ 2021            
        & P       & $1024\times3$                & 93.2            \\
        Centroid Transformers~\cite{wu2021centroid}% & Arxiv 2021         
        & P    & $1024\times3$                & 93.2            \\
        LFT-Net \cite{gao2022lft}                 %  & IEEE T-ITS 2022     
        & P, N    & $2048\times6$                & 93.2            \\
        3DMedPT~\cite{yu20213d}                    % & Arxiv 2021         
        & P       & $1024\times3$                & 93.4            \\
        3CROSSNet~\cite{han20223crossnet}          % & RA-L 2022          
        & P       & $1024\times3$                & 93.5            \\
        PatchFormer~\cite{cheng2021patchformer}    % & Arxiv 2021            
        & P, N    & $1024\times6$                & 93.6            \\
        Point Transformer~\cite{zhao2021point}     % & ICCV 2021            
        & P, N    & $1024\times6$                & 93.7            \\
        Point-BERT~\cite{yu2021pointbert}            %     & CVPR 2022       
        & P       & $8192\times3$                & 93.8            \\
        CAA~\cite{qiu2022geometric}                % & IEEE TMM 2022    
        & P       & $1024\times3$                & 93.8            \\
        PVT~\cite{zhang2021pvt}                    % & Arxiv 2022          
        & P, N    & $1024\times6$                & 94.0            \\

        \hline
    \end{tabular}
    \label{tab:classification}
    \vspace{-2ex}
\end{table}

\subsection{Classification \& Segmentation}
3D point cloud classification and segmentation are two fundamental yet challenging tasks, in which Transformers have played a key role. 
Classification can best reflect the ability of neural networks to extract salient features. Table.~\ref{tab:classification} shows the classification accuracy of different methods on the ModelNet40~\cite{wu20153d} dataset.
For fair comparisons, input data and input size are also shown. We report the Overall Accuracy (OA) as the evaluation metric, which is widely adopted.

From the table, we can see the recent proliferation of Transformer-based point cloud processing methods from 2020, when the Transformer architecture was first employed in image classification in the ViT paper~\cite{dosovitskiy2020image}.
Due to the strong ability of global information aggregating, Transformers rapidly achieved leading positions in this task.
Most 3D Transformers achieved a classification accuracy of around $93.0\%$. The newest PVT~\cite{zhang2021pvt} pushed the limit to $94.0\%$, which surpassed most non-Transformer algorithms of the same period. As an emerging technology, the success of the Transformer in point cloud classification demonstrates its great potential in the field of 3D point cloud processing.
We also presented the results of several state-of-the-art non-Transformer-based methods as reference.
As can be seen, the classification accuracy of the recent non-Transformer-based methods has exceeded $94.0\%$, and the highest one is $94.5\%$, achieved by PointMLP \cite{ma2022rethinking}. The various attention mechanisms used in Transformer methods are versatile and have great future potential for breakthroughs. We believe adapting innovations of general point cloud processing methods to Transformer methods can achieve state-of-the-art results. For example the Geometric Affine Module which resulted in PointMLP's impressive performance can be easily integrated into a Transformer-based network. I.e. PointMLP's performance is an indication of the Geometric Affine module's excellent performance, and not that MLPs are superior to Transformers as feature extraction backbones.   

%Therefore, it is hard to say which kind of algorithm is the best, and we believe that there will be new breakthroughs of 3D Transformers in the future.

%%%%%% part segmentation on ShapeNet %%%%%%%%%%%%%%%%%%%
\begin{table*}[htbp]
    \centering
    \caption{Comparative analysis between different point cloud Transformers in terms of pIoU on the ShapeNet part segmentation dataset. pIoU means part-average Intersection-over-Union.
        All results quoted were taken from the cited papers.}
    \setlength{\tabcolsep}{2pt}
    \begin{tabular}{l|c|cccccccccccccccc}
        \hline
        \textbf{Method}                % & {publication}  
        & {pIoU}        & \tabincell{c}{{air}-\\{plane}} & {bag}         & {cap}         & {car}         & {chair}       & \tabincell{c}{{ear}-\\{phone}} & {guitar}      & {knife}       & {lamp}        & {laptop}      & \tabincell{c}{motor-                                                                                 \\bike} & {mug} & {pistol} & {rocket} & \tabincell{c}{{skate}-\\{board}}& {table} \\
        \hline
        % PointNet~\cite{qi2017pointnet}             & CVPR 2017     & 83.7          & 83.4                                                 & 78.7          & 82.5          & 74.9          & 89.6          & 73.0                                                 & 91.5          & 85.9          & 80.8          & 95.3          & 65.2                 & 93.0          & 81.2          & 57.9          & 72.8          & 80.6          \\
        % PointNet++~\cite{qi2017pointnet++}         & NIPS 2017     & 85.1          & 82.4                                                 & 79.0          & 87.7          & 77.3          & 90.8          & 71.8                                                 & 91.0          & 85.9          & 83.7          & 95.3          & 71.6                 & 94.1          & 81.3          & 58.7          & 76.4          & 82.6          \\
        3DMedPT~\cite{yu20213d}                     %& Arxiv 2021 
        & 84.3          & 81.2& 86.0 &91.7& 79.6& 90.1& 81.2 &91.9 &88.5 &84.8 &96.0 &72.3 &95.8 &83.2 &64.6 &78.2 &83.8          \\
        ShapeContextNet~\cite{xie2018attentional}  %& CVPR 2018     
        & 84.6          & 83.8                                                 & 80.8          & 83.5          & 79.3          & 90.5          & 69.8                                                 & 91.7          & 86.5          & 82.9          & 96.0          & 69.2                 & 93.9          & 82.5          & 62.9          & 74.4          & 80.8          \\
        DT-Net~\cite{han2021dual}                   % & Arxiv 2021  
        & 85.6          & 83.0                                                 & 81.4          & 84.3          & 78.4          & 90.9          & 74.3                                                 & 91.0          & 87.3          & 84.7          & 95.6          & 69.0                 & 94.4          & 82.5          & 59.0          & 76.4          & 83.5          \\
        3CROSSNet~\cite{han20223crossnet}   %        & RA-L 2022  
        & 85.9          & 83.8                                                 & 84.9          & 86.1          & 79.8          & 91.2          & 70.3                                                 & 91.1          & 87.0          & 85.0          & 95.9          & 73.2                 & 94.9          & 83.2          & 56.2          & 76.7          & 83.0          \\
        CAA~\cite{qiu2022geometric}           %      & IEEE TMM 2022
        & 85.9          & 84.5                                                 & 82.2          & 86.8          & 78.9          & 91.1          & 74.5                                                 & 91.4          & 89.0          & 84.5         & 95.5          & 69.6                 & 94.2          & 83.4          & 57.8          & 75.5          & 83.5          \\
        PointASNL~\cite{yan2020pointasnl}          % & CVPR 2020  
        & 86.1          & 84.1                                                 & 84.7          & 87.9          & 79.7          & 92.2          & 73.7                                                 & 91.0          & 87.2          & 84.2          & 95.8          & 74.4                 & 95.2          & 81.0          & 63.0          & 76.3          & 83.2         \\
        LFT-Net \cite{gao2022lft}            %       & IEEE T-ITS 2022 
        & 86.2          & 83.0         & 83.9             & 90.9             & 79.4             & 93.1             & 71.4        & 92.5             & 88.6             & 85.7             & 95.9             & 69.3                    & 94.2             & 85.0             & 65.6             & 74.6             & 85.5             \\
        PCT~\cite{guo2021pct}                    %   & CVMJ 2021   
        & 86.4          & 85.0                                                 & 82.4          & 89.0          & 81.2          & 91.9          & 71.5                                                 & 91.3          & 88.1          & 86.3          & 95.8          & 64.6                 & 95.8          & 83.6          & 62.2          & 77.6          & 83.7          \\
        MLMSPT~\cite{han2021point}               %   & Arxiv 2021   
        & 86.4          & 84.4                                                 & 84.7          & 89.2          & 80.2          & 89.4          & 77.1                                                 & 92.3          & 87.5          & 85.3          & 96.7          & 71.6                 & 95.2          & 84.2          & 61.3          & 76.0          & 83.6          \\
        PatchFormer~\cite{cheng2021patchformer}   %  & Arxiv 2021 
        & 86.5          & -                                                    & -             & -             & -             & -             & -                                                    & -             & -             & -             & -             & -                    & -             & -             & -             & -             & -             \\
        % GAPNet~\cite{CHEN2021122}           & Neurocomputing 2021  &84.9	&84.0	&86.2	&88.8	&78.3	&90.7	&70.4	&91.3	&87.3	&82.8	&96.0	&68.7	&95.1	&82.0	&63.0	&74.8	&81.4 \\
        Point Transformer~\cite{zhao2021point}     % & ICCV 2021  
        & 86.6          & -                                                    & -             & -             & -             & -             & -                                                    & -             & -             & -             & -             & -                    & -             & -             & -             & -             & -             \\

        PVT~\cite{zhang2021pvt}                  %   & Arxiv 2022 
        & 86.6          & 85.3                                                 & 82.1          & 88.7          & 82.1          & 92.4          & 75.5                                                 & 91.0          & 88.9          & 85.6          & 95.4          & 76.2                 & 94.7          & 84.2          & 65.0          & 75.3          & 81.7 \\
        Stratified Transformer~\cite{lai2022stratified} %&  CVPR 2022  
        & 86.6          & -                                                    & -             & -             & -             & -             & -                                                    & -             & -             & -             & -             & -                    & -             & -             & -             & -             & -  \\

        \hline
    \end{tabular}
    \label{Tab.ShapeNet}
    \vspace{-1ex}
\end{table*}

%%%%%% semantic segmentation on S3DIS %%%%%%%%%%%%%%%%%%%
\begin{table*}[htbp]
\caption{Comparative analysis between different point cloud Transformers in terms of mIoU/mAcc/OA on the S3DIS Area 5 semantic segmentation dataset. mIoU means mean classwise Intersection over Union, mAcc means mean of classwise ACCuracy,
and OA means Overall pointwise Accuracy.
        All results quoted were taken from the cited papers.}
\label{tab:SemanticSegmentation}
\centering
\resizebox{\textwidth}{!}{
\begin{tabular}{c|c|c|c|cccccccccccccc}
\hline
\textbf{Method}  %& Publication 
&  OA &  mIoU &  mAcc & ceiling & floor & wall & beam & column & window & door & table & chair  & sofa & bookcase & board & clutter\\
\hline
ShapeContextNet \cite{xie2018attentional} %& CVPR 2018 
& 81.6 & 52.7 & -& -& -& -& -& -& -& -& -& -& -& -& -& -\\ 
% PointNet \cite{qi2017pointnet} & CVPR 2017 & - & 41.1 & 49.0 & 88.8 & 97.3 & 69.8 & 0.1 & 3.9 & 46.3 & 10.8 & 58.9 & 52.6 & 5.9 & 40.3 & 26.4 & 33.2 \\
PATs~\cite{yang2019modeling}    %& CVPR 2019
& - & 60.07 & 70.83 & 93.04 & 98.51 & 72.28 & 1.00 & 41.52 & 85.05 & 38.22 & 57.66 & 83.64 & 48.12 & 67.00 & 61.28 & 33.64 \\
PCT~\cite{guo2021pct} %& CVMJ 2021  
& - & 61.33 & 67.65& 92.54 &98.42 &80.62 &0.00 &19.37 &61.64 &48.00 &76.58 &85.20 &46.22 &67.71 &67.93 &52.29\\ 
PointASNL~\cite{yan2020pointasnl} %& CVPR 2020
& 87.7 & 62.6 & 68.5 & 94.3 & 98.4 & 79.1 & 0.0 & 26.7 & 55.2 & 66.2 & 83.3 & 86.8 & 47.6 & 68.3 & 56.4 & 52.1 \\    
MLMST~\cite{han2021point}    %& Arxiv 2021  
& -  &62.9 & - & 94.5 & 98.7 & 90.6 & 0.0 & 21.1 & 60.0 & 51.4 & 83.0 & 89.6 & 28.9 & 70.7 & 74.2 & 55.5\\
LFT-Net \cite{gao2022lft}                  % & IEEE T-ITS 2021 
& - & 65.2 & 76.2& 92.8& 96.1& 81.9& 0.0& 37.6& 70.3& 70.4& 73.2& 76.0& 40.9& 78.8& 71.0& 58.2\\
PVT~\cite{zhang2021pvt}                   % & Arxiv 2022 
& - & 67.30 & -& 91.18 &98.76 &86.23 &0.31 &34.21 &49.90 &61.45 &81.62 &89.85 &48.20 &79.96 &76.45 &54.67\\
EPT~\cite{park2022efficient} %& OpenReview 2022 
& - & 67.5 & 74.7& 91.5 &97.4 &86.0 &0.2 &40.4 &60.8 &66.7 &87.7 &79.6 &73.7 &58.6 &77.2 &57.3\\
PatchFormer~\cite{cheng2021patchformer}    %& Arxiv 2021 
& - & 68.1 & -& -& -& -& -& -& -& -& -& -& -& -& -& -\\
Point Transformer~\cite{zhao2021point}   %& ICCV 2021 
& 90.8 &70.4 & 76.5 &94.0 &98.5& 86.3 &0.0 &38.0 &63.4 &74.3 &89.1 &82.4 &74.3 &80.2 &76.0 &59.3 \\
% Cloud Transformers~\cite{mazur2021cloud}    & ICCV 2021 & - & 63.7 & -& -& -& -& -& -& -& -& -& -& -& -& -& -\\ 
Stratified Transformer~\cite{lai2022stratified} %& CVPR 2022 
& 91.5 & 72.0 & 78.1& -& -& -& -& -& -& -& -& -& -& -& -& -& -\\
\hline

\hline
\end{tabular}}
\end{table*}

For part segmentation, ShapeNet part segmentation dataset~\cite{yi2016scalable} results were used for comparison.
The commonly used part-average Intersection-over-Union was set as the performance metric.
As summarised in Table.~\ref{Tab.ShapeNet}, all the Transformer-based methods achieved a pIOU of around $86\%$, except for ShapeContextNet~\cite{xie2018attentional}, which was an early model published before 2019.
Note that Stratified Transformer~\cite{lai2022stratified} achieved the highest $86.6\%$ pIoU among all the comparative methods. It was also the best model in the task of semantic segmentation on the S3DIS semantic segmentation dataset~\cite{armeni20163d} (Table.~\ref{tab:SemanticSegmentation}).
% \subsection{Part segmentation}

% \subsection{Semantic segmentation}

\begin{table}[!t]
\caption{Comparative analysis between different point cloud Transformers in terms of AP on the ScanNetV2 and SUN RGB-D object detection datasets. AP means Average Precision. All results quoted were taken from the cited papers.
}
\centering
\begin{tabular}{@{}l|cccc@{}}
\hline
\textbf{Method} %& \textbf{Publication} 
&\multicolumn{2}{c}{\textbf{\scannet}} & \multicolumn{2}{c}{\textbf{\sunrgbd}}\\
& AP$_{25}$ & AP$_{50}$ & AP$_{25}$ & AP$_{50}$ \\
\hline
VoteNet~\cite{qi2019deep} %& ICCV 2019 
& 58.6 & 33.5 & 57.7 & - \\
3DETR~\cite{misra2021end}%& ICCV 2021 
& 62.7 & 37.5 & 56.8 & 30.1 \\
Pointformer~\cite{pan20213d}%& CVPR 2021 
& 64.1 & - & 61.1 & - \\
MLCVNet~\cite{xie2020mlcvnet} %& CVPR 2020 
& 64.5 & 41.4 & 59.8 & - \\
3DETR-m~\cite{misra2021end}%& ICCV 2021 
& 65.0 & 47.0 & 59.0 & 32.7 \\
GroupFree3D~\cite{liu2021group}%& ICCV 2021 
& 69.1 & 52.8 & 63.0 & 45.2 \\
\hline
\end{tabular}
\vspace{-0.1in}

\label{tab:detection}
\end{table}

\subsection{Object Detection}
The application of Transformers to 3D object detection from point clouds remains less explored research area.
There are only a few Transformer or Attention-based methods in recent literature.
A reason could be that object detection is more complicated than classification.
Table.~\ref{tab:detection} 
% and~\ref{scannet} 
summarises the performance of these Transformer-based networks on two public indoor scene datasets: SUN RGB-D~\cite{song2015sun} and ScanNetV2~\cite{dai2017scannet}.
VoteNet~\cite{qi2019deep} is also reported here as a reference, which is the pioneering work in 3D object detection.
In terms of $AP@25$ in the ScanNetV2 dataset, all the Transformer-based methods performed better than VoteNet.
Pointformer~\cite{pan20213d} and MLCVNet~\cite{xie2020mlcvnet} were based on VoteNet, and achieved similar performance.
Both of them utilized the self-attention mechanism in Transformers to enhance the feature representations.
Instead of leveraging the local voting strategy in the aforementioned two approaches, GroupFree3D~\cite{liu2021group} directly aggregated semantic information from all the points in the scene to extract the features of objects.
Its performance of $69.1\%$ demonstrated that aggregating features from all the elements by the self-attention mechanism is a more efficient way than the local voting strategy in VoteNet, MLCVNet, and Pointformer.
3DETR~\cite{misra2021end}, as the first end-to-end Transformer-based 3D object detector, achieved the second best detection performance, $65.0\%$, in the ScanNetV2 dataset.

%%%%%% 3D object detection %%%%%%%%%%%%%%%%%%%

%% file: 7_conclusion_future.tex
\section{Discussion and Conclusion}
\label{sec:7}

\subsection{Discussion}
As in 2D computer vision, Transformers also showed its potential in 3D point cloud processing.
From the perspective of the 3D tasks, Transformer-based methods mainly focused on high-level tasks, such as classification and segmentation.
We argue the reason is that Transformers are better at extracting global contextual information via capturing long-dependency relationships, which corresponds to the semantic information in high-level tasks.
On the other hand, low-level tasks, such as denoising and sampling, focus on exploring local geometric features.
From the perspective of performance, 3D Transformers improved the accuracy of the aforementioned tasks and surpassed most of the existing methods. However, as shown in Sec. \ref{sec:6}, for certain tasks, there is still a gap between them and the start-of-the-art non-Transformer-based methods. This is an indication that simply using Transformers as the backbone is not enough. Other innovative point cloud processing techniques must be employed.
Therefore, despite the rapid development of 3D Transformers, as an emerging technology, they still need further exploration and improvement. 

Based on the properties of Transformers and their successful applications in the 2D domain, we pointed out several potential future directions for 3D Transformers, hoping it will ignite the further development of this technology.

\subsubsection{Patch-wise Transformers}
As mentioned in Sec. \ref{subsec:2.2}, 3D Transformers can be divided into two groups: Point-wise Transformers and Channel-wise Transformers. Moreover, referring to the exploration of Transformers in 2D image processing  \cite{zhao2020exploring}, we are able to further divide Point-wise Transformers into Pair-wise Transformers and Patch-wise Transformers based on the operating form. The former calculates the attention weight for a feature vector by a corresponding pair of points, while the latter incorporates information from all points in a given patch.
% which is thus strictly more powerful than convolution. 
Specifically, the self-attention mechanism of pair-wise Transformers can be described as:
\begin{equation}
\label{pair-wise}
y_{i} = \sum_{j \in \Re_{i}}\alpha (x_{i},x_{j})\odot \beta (x_{j}),
\end{equation}
where $y_{i}$ is the output feature, $\Re_{i}$ is the operating scope of the self-attention module, $\odot$ is the Hadamard product, 
$\beta$ projects the feature $x_{j}$ to a new feature space by linear layers,
and $\alpha (x_{i},x_{j})$ is utilized to measure the relationship between $x_{i}$ and $x_{j}$, which can be decomposed as:
% \, \, \, \, $y_{i}$ is the output feature, 
% \, \, \, \, $\Re_{i}$ is the operating scope of the self-attention module, 
% \, \, \, \, $\beta$ projects the feature $x_{j}$ to a new feature space by linear layers, 
% \, \, \, \, $\alpha (x_{i},x_{j})$ is utilized to measure the relationship between $x_{i}$ and $x_{j}$, which can be decomposed as:
\begin{equation}
\alpha (x_{i},x_{j}) = \rho (\gamma (\delta (x_{i},x_{j}))),
\end{equation}
where $\rho$ is normalization function like softmax, 
$\gamma$ is a mapping function that ensures $\delta (x_{i},x_{j})$ has the same size as $\beta (x_{j})$, 
and $\delta$ is a relation function, the most common examples of which are:
% \, \, \, \, $\rho$ is normalization function like softmax, 
% \, \, \, \, $\gamma$ is a mapping function that ensures $\delta (x_{i},x_{j})$ has the same size as $\beta (x_{j})$, 
% \, \, \, \, $\delta$ is a relation function which can be expressed as:
\begin{equation}
\begin{aligned}
&\textit{Concatenation} : \delta(x_{i},x_{j}) = \left [ \varphi(x_{i}), \psi(x_{j}) \right ], \\
&\textit{Summation} : \delta(x_{i},x_{j}) = \varphi(x_{i}) + \psi(x_{j}), \\
&\textit{Subtraction} : \delta(x_{i},x_{j}) = \varphi(x_{i}) - \psi(x_{j}), \\
&\textit{Hadamard product} : \delta(x_{i},x_{j}) = \varphi(x_{i}) \odot \psi(x_{j}), \\
&\textit{Dot product}  : \delta(x_{i},x_{j}) = \varphi(x_{i})^{T}\psi(x_{j}), \\
\end{aligned}
\end{equation}
where the $\textit{Dot product}$ reduces this to a scalar attention operator, while the other forms reduce this to vector attention operators. The subtraction-form vector attention has been used in PT \cite{zhao2021point}.
From the Eq. \ref{pair-wise}, we can see that the attention weight $\alpha (x_{i},x_{j})$ is determined by a corresponding pair of point features $x_{i}$ and $x_{j}$. 
Pair-wise Transformers achieved compelling performance in the 2D image processing and were also commonly used in 3D point cloud processing. 
Nearly all algorithms in Sec. \ref{subsec:2.2.1} can be considered as pair-wise Transformers, where most of them used the $\textit{Dot product}$.

%Comment: unclear notation eq.11
Zhao et al. \cite{zhao2020exploring} also explored a family of patch-wise Transformers in image processing, whose self-attention mechanism can be expressed as:
\begin{equation}
\label{patch-wise}
y_{i} = \sum_{j \in \Re_{i}}\alpha (x_{\Re_{i}})_{j}\odot \beta (x_{j}),
\end{equation}
where $x_{\Re_{i}}$ is the patch of feature vectors in $\Re_{i}$, 
$\alpha$ transforms the $x_{\Re_{i}}$ to a new tensor with the same spatial dimensionality, 
and $\alpha (x_{\Re_{i}})_{j}$ is the $j$-th feature vector in this tensor. Similar to pair-wise Transformers, $\alpha (x_{\Re_{i}})$ can also be decomposed as:
% \, \, \, \, $x_{\Re_{i}}$ is the patch of feature vectors in $\Re_{i}$, 
% \, \, \, \, $\alpha$ transforms the $x_{\Re_{i}}$ to a new tensor with the same spatial dimensionality, 
% \, \, \, \,  $\alpha (x_{\Re_{i}})_{j}$ is the $j$-th feature vector in this tensor. Similar to pair-wise Transformers, $\alpha (x_{\Re_{i}})$ can also be decomposed as:
\begin{equation}
\alpha (x_{\Re_{i}}) = \rho (\gamma (\delta (x_{\Re_{i}}))),
\end{equation}
and $\delta$ can be expressed as three different forms \cite{zhao2020exploring}:
\begin{equation}
\begin{aligned}
&\textit{Concatenation} : \delta(x_{\Re_{i}}) = \left [ \varphi(x_{i}), [\psi(x_{j})]_{\forall j \in \Re_{i} } \right ], \\
&\textit{Star-product} : \delta(x_{\Re_{i}}) = [\varphi(x_{i})^{T}\psi(x_{j})]_{\forall j \in \Re_{i} }, \\
&\textit{Dot product}  : \delta(x_{\Re_{i}}) = [\varphi(x_{j})^{T}\psi(x_{k})]_{\forall j,k \in \Re_{i} }. \\
\end{aligned}
\end{equation}
By comparing Eq. \ref{pair-wise} and \ref{patch-wise}, we see that the latter aggregates all feature vectors in $\Re_{i}$ to generate the weight matrix that is applied to $\beta (x_{j})$, instead of merely utilizing a pair of features. 
In this way, patch-wise Transformers are able to enhance the connections among different feature vectors, and extract more robust short- and long-range dependencies. 
However, since the feature vectors are arranged in a particular order in $x_{\Re_{i}}$, patch-wise Transformers are not permutation-equivariant, which may have some negative effects on point cloud processing. 

Currently, there is little patch-wise Transformer research in the field of 3D point cloud processing. 
Considering the advantages of patch-wise Transformers and their outstanding performance in image processing, we believe that introducing patch-wise Transformers to point cloud processing is beneficial to performance improvement.

\subsubsection{Adaptive Set Abstraction}
PointNet++ \cite{qi2017pointnet++} proposed a Set Abstraction (SA) module to extract the semantic features of the point cloud hierarchically. It mainly utilized FPS and query ball grouping algorithms to achieve sampling point searching and local patch construction respectively. However, the sampling points generated by FPS tend to be evenly distributed in the original point cloud, while ignoring the geometric and semantic differences between different parts. For example, the tail of the aircraft is more geometrically complex and distinct than the fuselage. As such, the former needs more sampling points to be described. Moreover, query ball grouping focuses on searching the neighbor points only based on the Euclidean distance. However, it ignores the semantic feature differences among points, which makes it easy to group points with different semantic information into the same local patch. 
Therefore, developing an adaptive set abstraction is beneficial to improving the performance of 3D Transformers. Recently, there have been several Transformer-based methods in the 3D field exploring adaptive sampling \cite{wang2022lightn}. But few of them made full use of the rich short- and long-range dependencies generated by the self-attention mechanism. In the field of image processing, Deformable Attention Transformer (DAT) proposed in \cite{xia2022vision} generated the deformed sampling points by introducing an offset network. It achieved impressive results on comprehensive benchmarks with low computational footprint. It will be meaningful to present an adaptive sampling method based on the self-attention mechanism for the hierarchical Transformer. Additionally, inspired by the superpixel \cite{zhu2021learning} in the 2D field, we argue that it is feasible to utilize the attention map in 3D Transformers to obtain the ``superpoint" \cite{hui2021superpoint} for point cloud oversegmentation, converting point-level 3D data into neighborhood-level data. As such, this adaptive clustering technique can be used to replace the query ball grouping method.

\subsubsection{Self-supervised Transformer Pre-training}
Transformers have shown impressive performance on NLP and 2D image processing tasks. However, much of their success stems not only from their excellent scalability but also from large-scale self-supervised pre-training \cite{devlin2018bert}. Vision Transformer \cite{dosovitskiy2020image} performed a series of self-supervision experiments, and demonstrated the potential of the self-supervised Transformer. In the field of point cloud processing, despite the significant progress of supervised point cloud approaches, point cloud annotation is still a labor-intensive task. And the limited labeled dataset hinders the development of supervised approaches, especially in terms of the point cloud segmentation task. Recently, there have been a series of self-supervised approaches proposed to deal with these issues, such as Generative Adversarial Networks (GAN) \cite{goodfellow2014generative} in the 2D field, Auto-Encoders (AE) \cite{gadelha2018multiresolution, yang2017towards}, and Gaussian Mixture Models (GMM) \cite{achlioptas2018learning}. These methods used auto-encoders and generative models to realize self-supervised point cloud representation learning \cite{fu2022distillation}. Their satisfactory performances have demonstrated the effectiveness of the self-supervised point cloud approaches. However, few self-supervised Transformers have been currently applied to 3D point cloud processing. With the increasing availability of large-scale 3D point clouds, it is worthwhile to explore the self-supervised 3D Transformers for point cloud representation learning.

Overall, Transformers have only started to be applied to point cloud-related tasks. This research area has much space for innovations, especially by integrating breakthroughs from NLP and 2D computer vision. 

\subsection{Conclusion}
Transformer models have attracted widespread attention in the field of 3D point cloud processing, and achieved impressive results in various 3D tasks.
In this paper, we have comprehensively reviewed recent Transformer-based networks applied to point cloud-related tasks, such as point cloud classification, segmentation, object detection, registration, sampling, denoising, completion and other practical applications.
We first introduced the theory behind the Transformer architecture, and described the development and applications of 2D and 3D Transformers. 
Then we utilized three different taxonomies to categorize the current methods found in literature into multiple groups, and analyzed them from multiple perspectives. 
Additionally, we also described a series of self-attention variants that aimed to improve the performance and reduce the computational cost.
In terms of point cloud classification, segmentation and object detection, brief comparisons of the reviewed methods were provided in this paper.
Finally, we suggested three potential future research directions for the development of 3D Transformers.
We hope this survey gives researchers a comprehensive view of 3D Transformers, and drives their interest to further innovate the research in this field.

%% file: 0_main_TPAMI.bbl
\begin{thebibliography}{100}
\providecommand{\url}[1]{#1}
\csname url@samestyle\endcsname
\providecommand{\newblock}{\relax}
\providecommand{\bibinfo}[2]{#2}
\providecommand{\BIBentrySTDinterwordspacing}{\spaceskip=0pt\relax}
\providecommand{\BIBentryALTinterwordstretchfactor}{4}
\providecommand{\BIBentryALTinterwordspacing}{\spaceskip=\fontdimen2\font plus
\BIBentryALTinterwordstretchfactor\fontdimen3\font minus
  \fontdimen4\font\relax}
\providecommand{\BIBforeignlanguage}[2]{{%
\expandafter\ifx\csname l@#1\endcsname\relax
\typeout{** WARNING: IEEEtran.bst: No hyphenation pattern has been}%
\typeout{** loaded for the language `#1'. Using the pattern for}%
\typeout{** the default language instead.}%
\else
\language=\csname l@#1\endcsname
\fi
#2}}
\providecommand{\BIBdecl}{\relax}
\BIBdecl

\bibitem{han2022survey}
K.~Han \emph{et~al.}, ``A survey on vision transformer,'' \emph{IEEE Trans.
  Pattern Anal. Mach. Intell.}, 2022, doi:{
  \href{http://dx.doi.org/10.1109/TPAMI.2022.3152247}{10.1109/TPAMI.2022.3152247}}.

\bibitem{li2022contextual}
Y.~Li, T.~Yao, Y.~Pan, and T.~Mei, ``Contextual transformer networks for visual
  recognition,'' \emph{IEEE Trans. Pattern Anal. Mach. Intell.}, 2022, doi:{
  \href{http://dx.doi.org/10.1109/TPAMI.2022.3164083}{10.1109/TPAMI.2022.3164083}}.

\bibitem{xiao2022image}
J.~Xiao, X.~Fu, A.~Liu, F.~Wu, and Z.-J. Zha, ``Image de-raining transformer,''
  \emph{IEEE Trans. Pattern Anal. Mach. Intell.}, pp. 1--18, 2022, doi:{
  \href{http://dx.doi.org/10.1109/TPAMI.2022.3183612}{10.1109/TPAMI.2022.3183612}}.

\bibitem{qi2017pointnet++}
C.~R. Qi, L.~Yi, H.~Su, and L.~J. Guibas, ``Point{N}et++: Deep hierarchical
  feature learning on point sets in a metric space,'' in \emph{Proc. 31st Int.
  Conf. Neural Inf. Process. Syst.}, 2017, p. 5105–5114.

\bibitem{qi2017pointnet}
C.~R. Qi, H.~Su, K.~Mo, and L.~J. Guibas, ``Point{N}et: Deep learning on point
  sets for 3{D} classification and segmentation,'' in \emph{Proc. IEEE Conf.
  Comput. Vis. Pattern Recognit.}, 2017, pp. 77--85.

\bibitem{vaswani2017attention}
A.~Vaswani \emph{et~al.}, ``Attention is all you need,'' in \emph{Proc. 31st
  Int. Conf. Neural Inf. Process. Syst.}, 2017, pp. 6000--6010.

\bibitem{zhao2021point}
H.~Zhao, L.~Jiang, J.~Jia, P.~H. Torr, and V.~Koltun, ``Point transformer,'' in
  \emph{Proc. IEEE Int. Conf. Comput. Vis.}, 2021, pp. 16\,259--16\,268.

\bibitem{lu20223dctn}
\BIBentryALTinterwordspacing
D.~Lu, Q.~Xie, L.~Xu, and J.~Li, ``3{DCTN}: 3{D} convolution-transformer
  network for point cloud classification,'' \emph{arXiv:2203.00828}, 2022.
  [Online]. Available: \url{http://arxiv.org/abs/2203.00828}
\BIBentrySTDinterwordspacing

\bibitem{tancik2020fourier}
M.~Tancik \emph{et~al.}, ``Fourier features let networks learn high frequency
  functions in low dimensional domains,'' in \emph{Proc. Adv. Neural Inf.
  Process. Syst.}, 2020, pp. 7537--7547.

\bibitem{feng2020point}
M.~Feng, L.~Zhang, X.~Lin, S.~Z. Gilani, and A.~Mian, ``Point attention network
  for semantic segmentation of 3{D} point clouds,'' \emph{Pattern Recognit.},
  vol. 107, p. 107446, 2020, doi:{
  \href{http://dx.doi.org/10.1016/j.patcog.2020.107446}{10.1016/j.patcog.2020.107446}}.

\bibitem{xie2018attentional}
S.~Xie, S.~Liu, Z.~Chen, and Z.~Tu, ``Attentional shapecontextnet for point
  cloud recognition,'' in \emph{Proc. IEEE Conf. Comput. Vis. Pattern
  Recognit.}, 2018, pp. 4606--4615.

\bibitem{guo2021pct}
M.-H. Guo, J.-X. Cai, Z.-N. Liu, T.-J. Mu, R.~R. Martin, and S.-M. Hu, ``P{CT}:
  Point cloud transformer,'' \emph{Comput. Vis. Media.}, vol.~7, no.~2, pp.
  187--199, 2021.

\bibitem{wang2019dynamic}
Y.~Wang, Y.~Sun, Z.~Liu, S.~E. Sarma, M.~M. Bronstein, and J.~M. Solomon,
  ``Dynamic graph {CNN} for learning on point clouds,'' \emph{ACM Trans.
  Graph.}, vol.~38, no.~5, pp. 1--12, 2019.

\bibitem{wang2021max}
H.~Wang, Y.~Zhu, H.~Adam, A.~Yuille, and L.-C. Chen, ``Max-deeplab: End-to-end
  panoptic segmentation with mask transformers,'' in \emph{Proc. IEEE Conf.
  Comput. Vis. Pattern Recognit.}, 2021, pp. 5463--5474.

\bibitem{carion2020end}
N.~Carion \emph{et~al.}, ``End-to-end object detection with transformers,'' in
  \emph{Proc. Eur. Conf. Comput. Vis.}, vol. 12346, 2020, pp. 213--229.

\bibitem{chen2021transformer}
X.~Chen, B.~Yan, J.~Zhu, D.~Wang, X.~Yang, and H.~Lu, ``Transformer tracking,''
  in \emph{Proc. IEEE Conf. Comput. Vis. Pattern Recognit.}, 2021, pp.
  8126--8135.

\bibitem{dosovitskiy2020image}
A.~Dosovitskiy \emph{et~al.}, ``An image is worth 16x16 words: Transformers for
  image recognition at scale,'' in \emph{Proc. Int. Conf. Learn. Represent.},
  2020, pp. 1--12.

\bibitem{wu2020visual}
\BIBentryALTinterwordspacing
B.~Wu \emph{et~al.}, ``Visual transformers: Token-based image representation
  and processing for computer vision,'' \emph{arXiv:2006.03677}, 2020.
  [Online]. Available: \url{http://arxiv.org/abs/2006.03677}
\BIBentrySTDinterwordspacing

\bibitem{wang2021pyramid}
W.~Wang \emph{et~al.}, ``Pyramid vision transformer: A versatile backbone for
  dense prediction without convolutions,'' in \emph{Proc. IEEE Int. Conf.
  Comput. Vis.}, 2021, pp. 548--558.

\bibitem{wu2021cvt}
H.~Wu \emph{et~al.}, ``Cv{T}: Introducing convolutions to vision
  transformers,'' in \emph{Proc. IEEE Int. Conf. Comput. Vis.}, 2021, pp.
  22--31.

\bibitem{liu2021swin}
Z.~Liu, Y.~Lin, Y.~Cao, H.~Hu, Y.~Wei, Z.~Zhang, S.~Lin, and B.~Guo, ``Swin
  transformer: Hierarchical vision transformer using shifted windows,'' in
  \emph{Proc. IEEE Int. Conf. Comput. Vis.}, 2021, pp. 9992--10\,002.

\bibitem{xie2021segformer}
E.~Xie and othersg, ``Segformer: Simple and efficient design for semantic
  segmentation with transformers,'' in \emph{Proc. Adv. Neural Inf. Process.
  Syst.}, 2021, pp. 12\,077--12\,090.

\bibitem{cheng2021per}
B.~Cheng, A.~Schwing, and A.~Kirillov, ``Per-pixel classification is not all
  you need for semantic segmentation,'' in \emph{Proc. Adv. Neural Inf.
  Process. Syst.}, 2021, pp. 17\,864--17\,875.

\bibitem{wang2021end}
Y.~Wang \emph{et~al.}, ``End-to-end video instance segmentation with
  transformers,'' in \emph{Proc. IEEE Conf. Comput. Vis. Pattern Recognit.},
  2021, pp. 8741--8750.

\bibitem{yao2021efficient}
\BIBentryALTinterwordspacing
Z.~Yao, J.~Ai, B.~Li, and C.~Zhang, ``Efficient {DETR}: improving end-to-end
  object detector with dense prior,'' \emph{arXiv:2104.01318}, 2021. [Online].
  Available: \url{http://arxiv.org/abs/2104.01318}
\BIBentrySTDinterwordspacing

\bibitem{yang2021focal}
\BIBentryALTinterwordspacing
J.~Yang \emph{et~al.}, ``Focal self-attention for local-global interactions in
  vision transformers,'' \emph{arXiv:2107.00641}, 2021. [Online]. Available:
  \url{http://arxiv.org/abs/2107.00641}
\BIBentrySTDinterwordspacing

\bibitem{chen2021empirical}
X.~Chen, S.~Xie, and K.~He, ``An empirical study of training self-supervised
  vision transformers,'' in \emph{Proc. IEEE Int. Conf. Comput. Vis.}, 2021,
  pp. 9640--9649.

\bibitem{liu2021survey}
\BIBentryALTinterwordspacing
Y.~Liu \emph{et~al.}, ``A survey of visual transformers,''
  \emph{arXiv:2111.06091}, 2021. [Online]. Available:
  \url{http://arxiv.org/abs/2111.06091}
\BIBentrySTDinterwordspacing

\bibitem{khan2021transformers}
S.~Khan, M.~Naseer, M.~Hayat, S.~W. Zamir, F.~S. Khan, and M.~Shah,
  ``Transformers in vision: A survey,'' \emph{ACM Computing Surveys (CSUR)},
  2021.

\bibitem{xu2022multimodal}
\BIBentryALTinterwordspacing
P.~Xu, X.~Zhu, and D.~A. Clifton, ``Multimodal learning with transformers: A
  survey,'' \emph{arXiv:2206.06488}, 2022. [Online]. Available:
  \url{http://arxiv.org/abs/2206.06488}
\BIBentrySTDinterwordspacing

\bibitem{han20223crossnet}
X.-F. Han, Z.-Y. He, J.~Chen, and G.-Q. Xiao, ``3{CROSSNet}: Cross-level
  cross-scale cross-attention network for point cloud representation,''
  \emph{IEEE Robotics Autom. Lett.}, vol.~7, no.~2, pp. 3718--3725, 2022.

\bibitem{yan2020pointasnl}
X.~Yan, C.~Zheng, Z.~Li, S.~Wang, and S.~Cui, ``Point{ASNL}: Robust point
  clouds processing using nonlocal neural networks with adaptive sampling,'' in
  \emph{Proc. IEEE Conf. Comput. Vis. Pattern Recognit.}, 2020, pp. 5589--5598.

\bibitem{yu2021pointbert}
X.~Yu, L.~Tang, Y.~Rao, T.~Huang, J.~Zhou, and J.~Lu, ``Point-{BERT}:
  Pre-training 3{D} point cloud transformers with masked point modeling,'' in
  \emph{Proc. IEEE Conf. Comput. Vis. Pattern Recognit.}, 2022, pp.
  19\,313--19\,322.

\bibitem{gao2022lft}
Y.~Gao, X.~Liu, J.~Li, Z.~Fang, X.~Jiang, and K.~M.~S. Huq, ``{LFT-N}et: Local
  feature transformer network for point clouds analysis,'' \emph{IEEE Trans.
  Intell. Transport. Syst.}, 2022, doi:{
  \href{http://dx.doi.org/10.1109/TITS.2022.3140355}{10.1109/TITS.2022.3140355}}.

\bibitem{pan20213d}
X.~Pan, Z.~Xia, S.~Song, L.~E. Li, and G.~Huang, ``3{D} object detection with
  pointformer,'' in \emph{Proc. IEEE Conf. Comput. Vis. Pattern Recognit.},
  2021, pp. 7463--7472.

\bibitem{xu2022tdnet}
X.~Xu, G.~Geng, X.~Cao, K.~Li, and M.~Zhou, ``{TDNet}: transformer-based
  network for point cloud denoising,'' \emph{Appl. Opt.}, vol.~61, no.~6, pp.
  C80--C88, 2022.

\bibitem{yu20213d}
\BIBentryALTinterwordspacing
J.~Yu \emph{et~al.}, ``3{D} medical point transformer: Introducing convolution
  to attention networks for medical point cloud analysis,''
  \emph{arXiv:2112.04863}, 2021. [Online]. Available:
  \url{http://arxiv.org/abs/2112.04863}
\BIBentrySTDinterwordspacing

\bibitem{qiu2021pu}
\BIBentryALTinterwordspacing
S.~Qiu, S.~Anwar, and N.~Barnes, ``P{U}-{T}ransformer: Point cloud upsampling
  transformer,'' \emph{arXiv:2111.12242}, 2021. [Online]. Available:
  \url{http://arxiv.org/abs/2111.12242}
\BIBentrySTDinterwordspacing

\bibitem{han2021dual}
\BIBentryALTinterwordspacing
X.-F. Han, Y.-F. Jin, H.-X. Cheng, and G.-Q. Xiao, ``Dual transformer for point
  cloud analysis,'' \emph{arXiv:2104.13044}, 2021. [Online]. Available:
  \url{http://arxiv.org/abs/2104.13044}
\BIBentrySTDinterwordspacing

\bibitem{qiu2022geometric}
S.~Qiu, S.~Anwar, and N.~Barnes, ``Geometric back-projection network for point
  cloud classification,'' \emph{IEEE Trans. Multimedia}, vol.~24, pp.
  1943--1955, 2022.

\bibitem{xu2021adaptive}
\BIBentryALTinterwordspacing
G.~Xu, H.~Cao, J.~Wan, K.~Xu, Y.~Ma, and C.~Zhang, ``Adaptive channel encoding
  transformer for point cloud analysis,'' \emph{arXiv:2112.02507}, 2021.
  [Online]. Available: \url{http://arxiv.org/abs/2112.02507}
\BIBentrySTDinterwordspacing

\bibitem{wu2021centroid}
\BIBentryALTinterwordspacing
L.~Wu, X.~Liu, and Q.~Liu, ``Centroid transformers: Learning to abstract with
  attention,'' \emph{arXiv:2102.08606}, 2021. [Online]. Available:
  \url{http://arxiv.org/abs/2102.08606}
\BIBentrySTDinterwordspacing

\bibitem{wang2022lightn}
\BIBentryALTinterwordspacing
X.~Wang, Y.~Jin, Y.~Cen, T.~Wang, B.~Tang, and Y.~Li, ``{LighTN}: Light-weight
  transformer network for performance-overhead tradeoff in point cloud
  downsampling,'' \emph{arXiv:2202.06263}, 2022. [Online]. Available:
  \url{http://arxiv.org/abs/2202.06263}
\BIBentrySTDinterwordspacing

\bibitem{yang2019modeling}
J.~Yang \emph{et~al.}, ``Modeling point clouds with self-attention and gumbel
  subset sampling,'' in \emph{Proc. IEEE Conf. Comput. Vis. Pattern Recognit.},
  2019, pp. 3323--3332.

\bibitem{zhang2021pvt}
\BIBentryALTinterwordspacing
C.~Zhang, H.~Wan, S.~Liu, X.~Shen, and Z.~Wu, ``P{VT}: Point-voxel transformer
  for 3{D} deep learning,'' \emph{arXiv:2108.06076}, 2021. [Online]. Available:
  \url{http://arxiv.org/abs/2108.06076}
\BIBentrySTDinterwordspacing

\bibitem{mao2021voxel}
J.~Mao \emph{et~al.}, ``Voxel transformer for 3{D} object detection,'' in
  \emph{Proc. IEEE Int. Conf. Comput. Vis.}, 2021, pp. 3164--3173.

\bibitem{he2022voxset}
S.~L. Chenhang~He, Ruihuang~Li and L.~Zhang, ``Voxel set transformer: A
  set-to-set approach to 3{D} object detection from point clouds,'' in
  \emph{Proc. IEEE Conf. Comput. Vis. Pattern Recognit.}, 2022, pp. 8417--8427.

\bibitem{fan2021svt}
Z.~Fan, Z.~Song, H.~Liu, Z.~Lu, J.~He, and X.~Du, ``{SVT-N}et: Super
  light-weight sparse voxel transformer for large scale place recognition,'' in
  \emph{Proc. AAAI Conf. Artif. Intell.}, 2022, pp. 551--560.

\bibitem{park2022efficient}
\BIBentryALTinterwordspacing
C.~Park, Y.~Jeong, M.~Cho, and J.~Park, ``Efficient point transformer for
  large-scale 3{D} scene understanding,'' 2022. [Online]. Available:
  \url{https://openreview.net/forum?id=3SUToIxuIT3}
\BIBentrySTDinterwordspacing

\bibitem{lin2021pctma}
J.~Lin, M.~Rickert, A.~Perzylo, and A.~Knoll, ``{PCTMA-N}et: Point cloud
  transformer with morphing atlas-based point generation network for dense
  point cloud completion,'' in \emph{Proc. IEEE/RSJ Int. Conf. Intell. Robots
  Syst.}, 2021, pp. 5657--5663.

\bibitem{lai2022stratified}
X.~Lai \emph{et~al.}, ``Stratified transformer for 3{D} point cloud
  segmentation,'' in \emph{Proc. IEEE Conf. Comput. Vis. Pattern Recognit.},
  2022, pp. 8500--8509.

\bibitem{hui2021pyramid}
L.~Hui, H.~Yang, M.~Cheng, J.~Xie, and J.~Yang, ``Pyramid point cloud
  transformer for large-scale place recognition,'' in \emph{Proc. IEEE Int.
  Conf. Comput. Vis.}, 2021, pp. 6098--6107.

\bibitem{xie2020mlcvnet}
Q.~Xie \emph{et~al.}, ``{MLCVN}et: Multi-level context votenet for 3{D} object
  detection,'' in \emph{Proc. IEEE Conf. Comput. Vis. Pattern Recognit.}, 2020,
  pp. 10\,447--10\,456.

\bibitem{liu2021group}
Z.~Liu, Z.~Zhang, Y.~Cao, H.~Hu, and X.~Tong, ``Group-free 3{D} object
  detection via transformers,'' in \emph{Proc. IEEE Int. Conf. Comput. Vis.},
  2021, pp. 2949--2958.

\bibitem{misra2021end}
I.~Misra, R.~Girdhar, and A.~Joulin, ``An end-to-end transformer model for 3{D}
  object detection,'' in \emph{Proc. IEEE Int. Conf. Comput. Vis.}, 2021, pp.
  2906--2917.

\bibitem{cui20213d}
Y.~Cui, Z.~Fang, J.~Shan, Z.~Gu, and S.~Zhou, ``3{D} object tracking with
  transformer,'' \emph{Proc. Brit. Mach. Vis. Conf.}, p. 317, 2021.

\bibitem{zhou2021pttr}
C.~Zhou \emph{et~al.}, ``{PTTR}: Relational 3{D} point cloud object tracking
  with transformer,'' in \emph{Proc. IEEE Conf. Comput. Vis. Pattern
  Recognit.}, 2022, pp. 8531--8540.

\bibitem{jiayao2022real}
S.~Jiayao, S.~Zhou, Y.~Cui, and Z.~Fang, ``Real-time 3{D} single object
  tracking with transformer,'' \emph{IEEE Trans. Multimedia}, 2022, doi:{
  \href{http://dx.doi.org/10.1109/TMM.2022.3146714}{10.1109/TMM.2022.3146714}}.

\bibitem{wang2019deep}
Y.~Wang and J.~M. Solomon, ``Deep closest point: Learning representations for
  point cloud registration,'' in \emph{Proc. IEEE Int. Conf. Comput. Vis.},
  2019, pp. 3523--3532.

\bibitem{wang2022storm}
Y.~Wang, C.~Yan, Y.~Feng, S.~Du, Q.~Dai, and Y.~Gao, ``{STORM}: Structure-based
  overlap matching for partial point cloud registration,'' \emph{IEEE Trans.
  Pattern Anal. Mach. Intell.}, 2022, doi:{
  \href{http://dx.doi.org/10.1109/TPAMI.2022.3148308}{10.1109/TPAMI.2022.3148308}}.

\bibitem{fischer2021stickypillars}
K.~Fischer \emph{et~al.}, ``Sticky{P}illars: Robust and efficient feature
  matching on point clouds using graph neural networks,'' in \emph{Proc. IEEE
  Conf. Comput. Vis. Pattern Recognit.}, 2021, pp. 313--323.

\bibitem{fu2021robust}
K.~Fu, S.~Liu, X.~Luo, and M.~Wang, ``Robust point cloud registration framework
  based on deep graph matching,'' in \emph{Proc. IEEE Conf. Comput. Vis.
  Pattern Recognit.}, 2021, pp. 8893--8902.

\bibitem{chen2021full}
\BIBentryALTinterwordspacing
G.~Chen, M.~Wang, Y.~Yue, Q.~Zhang, and L.~Yuan, ``Full transformer framework
  for robust point cloud registration with deep information interaction,''
  \emph{arXiv:2112.09385}, 2021. [Online]. Available:
  \url{http://arxiv.org/abs/2112.09385}
\BIBentrySTDinterwordspacing

\bibitem{fan2021point}
H.~Fan, Y.~Yang, and M.~Kankanhalli, ``Point 4{D} transformer networks for
  spatio-temporal modeling in point cloud videos,'' in \emph{Proc. IEEE Conf.
  Comput. Vis. Pattern Recognit.}, 2021, pp. 14\,204--14\,213.

\bibitem{gao2022reflective}
R.~Gao, M.~Li, S.-J. Yang, and K.~Cho, ``Reflective noise filtering of
  large-scale point cloud using transformer,'' \emph{Remote Sens.}, vol.~14,
  no.~3, p. 577, 2022.

\bibitem{yu2021pointr}
X.~Yu, Y.~Rao, Z.~Wang, Z.~Liu, J.~Lu, and J.~Zhou, ``Poin{T}r: Diverse point
  cloud completion with geometry-aware transformers,'' in \emph{Proc. IEEE Int.
  Conf. Comput. Vis.}, 2021, pp. 12\,498--12\,507.

\bibitem{xiang2021snowflakenet}
P.~Xiang \emph{et~al.}, ``Snowflake{N}et: Point cloud completion by snowflake
  point deconvolution with skip-transformer,'' in \emph{Proc. IEEE Int. Conf.
  Comput. Vis.}, 2021, pp. 5499--5509.

\bibitem{yan2022shapeformer}
\BIBentryALTinterwordspacing
X.~Yan, L.~Lin, N.~J. Mitra, D.~Lischinski, D.~Cohen-Or, and H.~Huang,
  ``Shape{F}ormer: Transformer-based shape completion via sparse
  representation,'' \emph{arXiv:2201.10326}, 2022. [Online]. Available:
  \url{http://arxiv.org/abs/2201.10326}
\BIBentrySTDinterwordspacing

\bibitem{sheng2021improving}
H.~Sheng, S.~Cai, Y.~Liu, B.~Deng, J.~Huang, X.-S. Hua, and M.-J. Zhao,
  ``Improving 3{D} object detection with channel-wise transformer,'' in
  \emph{Proc. IEEE Int. Conf. Comput. Vis.}, 2021, pp. 2743--2752.

\bibitem{wei2022spatial}
Y.~Wei, H.~Liu, T.~Xie, Q.~Ke, and Y.~Guo, ``Spatial-temporal transformer for
  3{D} point cloud sequences,'' in \emph{Proc. IEEE Winter Conf. Appl. Comput.
  Vis.}, 2022, pp. 1171--1180.

\bibitem{qin2022geometric}
Z.~Qin, H.~Yu, C.~Wang, Y.~Guo, Y.~Peng, and K.~Xu, ``Geometric transformer for
  fast and robust point cloud registration,'' in \emph{Proc. IEEE Conf. Comput.
  Vis. Pattern Recognit.}, 2022, pp. 11\,143--11\,152.

\bibitem{yew2022regtr}
Z.~J. Yew and G.~h. Lee, ``{REGTR}: End-to-end point cloud correspondences with
  transformers,'' in \emph{Proc. IEEE Conf. Comput. Vis. Pattern Recognit.},
  2022, pp. 6677--6686.

\bibitem{chen2021transsc}
\BIBentryALTinterwordspacing
W.~Chen, H.~Liang, Z.~Chen, F.~Sun, and J.~Zhang, ``Trans{SC}:
  Transformer-based shape completion for grasp evaluation,''
  \emph{arXiv:2107.00511}, 2021. [Online]. Available:
  \url{http://arxiv.org/abs/2107.00511}
\BIBentrySTDinterwordspacing

\bibitem{liu2022point}
X.~Liu, G.~Xu, K.~Xu, J.~Wan, and Y.~Ma, ``Point cloud completion by dynamic
  transformer with adaptive neighbourhood feature fusion,'' \emph{IET Comput.
  Vis.}, 2022, doi:{
  \href{https://doi.org/10.1049/cvi2.12098}{10.1049/cvi2.12098}}.

\bibitem{zhou2022sewer}
Y.~Zhou, A.~Ji, and L.~Zhang, ``Sewer defect detection from 3{D} point clouds
  using a transformer-based deep learning model,'' \emph{Autom. Constr.}, vol.
  136, p. 104163, 2022.

\bibitem{prakash2021multi}
A.~Prakash, K.~Chitta, and A.~Geiger, ``Multi-modal fusion transformer for
  end-to-end autonomous driving,'' in \emph{Proc. IEEE Conf. Comput. Vis.
  Pattern Recognit.}, 2021, pp. 7077--7087.

\bibitem{yuan2021temporal}
Z.~Yuan, X.~Song, L.~Bai, Z.~Wang, and W.~Ouyang, ``Temporal-channel
  transformer for 3{D} lidar-based video object detection for autonomous
  driving,'' \emph{IEEE Trans. Circuits Syst. Video Technol.}, vol.~32, no.~4,
  pp. 2068--2078, 2021.

\bibitem{qiu2021investigating}
S.~Qiu, Y.~Wu, S.~Anwar, and C.~Li, ``Investigating attention mechanism in 3{D}
  point cloud object detection,'' in \emph{3DV}, 2021, pp. 403--412.

\bibitem{song2015sun}
S.~Song, S.~P. Lichtenberg, and J.~Xiao, ``{SUN RGB-D}: A {RGB-D} scene
  understanding benchmark suite,'' in \emph{Proc. IEEE Conf. Comput. Vis.
  Pattern Recognit.}, 2015, pp. 567--576.

\bibitem{dai2017scannet}
A.~Dai \emph{et~al.}, ``Scan{N}et: Richly-annotated 3{D} reconstructions of
  indoor scenes,'' in \emph{Proc. IEEE Conf. Comput. Vis. Pattern Recognit.},
  2017, pp. 5828--5839.

\bibitem{gao2021multi}
X.-Y. Gao, Y.-Z. Wang, C.-X. Zhang, and J.-Q. Lu, ``Multi-head self-attention
  for 3{D} point cloud classification,'' \emph{IEEE Access}, vol.~9, pp.
  18\,137--18\,147, 2021.

\bibitem{bruna2013spectral}
J.~Bruna, W.~Zaremba, A.~Szlam, and Y.~LeCun, ``Spectral networks and locally
  connected networks on graphs,'' in \emph{Proc. Int. Conf. Learn. Represent.},
  2014.

\bibitem{devlin2018bert}
J.~Devlin, M.-W. Chang, K.~Lee, and K.~Toutanova, ``{BERT:} pre-training of
  deep bidirectional transformers for language understanding,'' in
  \emph{NAACL-HLT}, 2019, pp. 4171--4186.

\bibitem{rolfe2016discrete}
\BIBentryALTinterwordspacing
J.~T. Rolfe, ``Discrete variational autoencoders,'' \emph{arXiv:1609.02200},
  2016. [Online]. Available: \url{http://arxiv.org/abs/1609.02200}
\BIBentrySTDinterwordspacing

\bibitem{wang2022local}
Z.~Wang, Y.~Wang, L.~An, J.~Liu, and H.~Liu, ``Local transformer network on
  3{D} point cloud semantic segmentation,'' \emph{Information}, vol.~13, no.~4,
  p. 198, 2022.

\bibitem{liu2022group}
S.~Liu, K.~Fu, M.~Wang, and Z.~Song, ``Group-in-group relation-based
  transformer for 3{D} point cloud learning,'' \emph{Remote Sens.}, vol.~14,
  no.~7, p. 1563, 2022.

\bibitem{zhao2020exploring}
H.~Zhao, J.~Jia, and V.~Koltun, ``Exploring self-attention for image
  recognition,'' in \emph{Proc. IEEE Conf. Comput. Vis. Pattern Recognit.},
  2020, pp. 10\,076--10\,085.

\bibitem{thomas2019kpconv}
H.~Thomas \emph{et~al.}, ``K{P}conv: Flexible and deformable convolution for
  point clouds,'' in \emph{Proc. IEEE Int. Conf. Comput. Vis.}, 2019, pp.
  6411--6420.

\bibitem{cheng2021patchformer}
\BIBentryALTinterwordspacing
Z.~Cheng, H.~Wan, X.~Shen, and Z.~Wu, ``Patchformer: A versatile 3{D}
  transformer based on patch attention,'' \emph{arXiv:2111.00207}, 2021.
  [Online]. Available: \url{http://arxiv.org/abs/2111.00207}
\BIBentrySTDinterwordspacing

\bibitem{mehta2020delight}
\BIBentryALTinterwordspacing
S.~Mehta, M.~Ghazvininejad, S.~Iyer, L.~Zettlemoyer, and H.~Hajishirzi,
  ``Delight: Deep and light-weight transformer,'' \emph{arXiv:2008.00623},
  2020. [Online]. Available: \url{http://arxiv.org/abs/2008.00623}
\BIBentrySTDinterwordspacing

\bibitem{xu2021voxel}
Y.~Xu, X.~Tong, and U.~Stilla, ``Voxel-based representation of 3{D} point
  clouds: Methods, applications, and its potential use in the construction
  industry,'' \emph{Autom. Constr.}, vol. 126, p. 103675, 2021.

\bibitem{graham20183d}
B.~Graham, M.~Engelcke, and L.~Van Der~Maaten, ``3{D} semantic segmentation
  with submanifold sparse convolutional networks,'' in \emph{Proc. IEEE Conf.
  Comput. Vis. Pattern Recognit.}, 2018, pp. 9224--9232.

\bibitem{choy20194d}
C.~Choy, J.~Gwak, and S.~Savarese, ``4{D} spatio-temporal convnets: Minkowski
  convolutional neural networks,'' in \emph{Proc. IEEE Conf. Comput. Vis.
  Pattern Recognit.}, 2019, pp. 3075--3084.

\bibitem{lee2019set}
J.~Lee \emph{et~al.}, ``Set transformer: A framework for attention-based
  permutation-invariant neural networks,'' in \emph{Proc. Int. Conf. Mach.
  Learn.}, 2019, pp. 3744--3753.

\bibitem{han2021point}
\BIBentryALTinterwordspacing
X.-F. Han, Y.-J. Kuang, and G.-Q. Xiao, ``Point cloud learning with
  transformer,'' \emph{arXiv:2104.13636}, 2021. [Online]. Available:
  \url{http://arxiv.org/abs/2104.13636}
\BIBentrySTDinterwordspacing

\bibitem{fu2022distillation}
\BIBentryALTinterwordspacing
K.~Fu, P.~Gao, R.~Zhang, H.~Li, Y.~Qiao, and M.~Wang, ``Distillation with
  contrast is all you need for self-supervised point cloud representation
  learning,'' \emph{arXiv:2202.04241}, 2022. [Online]. Available:
  \url{http://arxiv.org/abs/2202.04241}
\BIBentrySTDinterwordspacing

\bibitem{yang2022mil}
C.-K. Yang, J.-J. Wu, K.-S. Chen, Y.-Y. Chuang, and Y.-Y. Lin, ``An
  {MIL-D}erived transformer for weakly supervised point cloud segmentation,''
  in \emph{Proc. IEEE Conf. Comput. Vis. Pattern Recognit.}, 2022, pp.
  11\,830--11\,839.

\bibitem{park2022fast}
C.~Park, Y.~Jeong, M.~Cho, and J.~Park, ``Fast point transformer,'' in
  \emph{Proc. IEEE Conf. Comput. Vis. Pattern Recognit.}, 2022, pp.
  16\,949--16\,958.

\bibitem{zhang2021u}
J.~Zhang, X.~Li, X.~Zhao, Y.~Ge, and Z.~Zhang, ``U-shaped network based on
  transformer for 3{D} point clouds semantic segmentation,'' in \emph{ICVIP},
  2021, pp. 170--176.

\bibitem{chen2022pq}
X.~Chen, H.~Zhao, G.~Zhou, and Y.-Q. Zhang, ``{PQ}-transformer: Jointly parsing
  3{D} objects and layouts from point clouds,'' \emph{IEEE Robot. Autom.
  Lett.}, vol.~7, no.~2, pp. 2519--2526, 2022.

\bibitem{zhang2022cat}
Y.~Zhang, J.~Chen, and D.~Huang, ``{CAT-D}et: Contrastively augmented
  transformer for multi-modal 3{D} object detection,'' \emph{Proc. IEEE Conf.
  Comput. Vis. Pattern Recognit.}, 2022.

\bibitem{wangbridged}
Y.~Wang \emph{et~al.}, ``Bridged transformer for vision and point cloud 3{D}
  object detection,'' \emph{Proc. IEEE Conf. Comput. Vis. Pattern Recognit.},
  pp. 12\,114--12\,123, 2022.

\bibitem{trappolini2021shape}
G.~Trappolini, L.~Cosmo, L.~Moschella, R.~Marin, S.~Melzi, and E.~Rodol{\`a},
  ``Shape registration in the time of transformers,'' in \emph{Proc. Adv.
  Neural Inf. Process. Syst.}, 2021, pp. 5731--5744.

\bibitem{krizhevsky2012imagenet}
A.~Krizhevsky, I.~Sutskever, and G.~E. Hinton, ``Imagenet classification with
  deep convolutional neural networks,'' \emph{Proc. Adv. Neural Inf. Process.
  Syst.}, pp. 1106--1114, 2012.

\bibitem{simonyan2014very}
\BIBentryALTinterwordspacing
K.~Simonyan and A.~Zisserman, ``Very deep convolutional networks for
  large-scale image recognition,'' \emph{arXiv:1409.1556}, 2014. [Online].
  Available: \url{http://arxiv.org/abs/1409.1556}
\BIBentrySTDinterwordspacing

\bibitem{he2016deep}
K.~He, X.~Zhang, S.~Ren, and J.~Sun, ``Deep residual learning for image
  recognition,'' in \emph{Proc. IEEE Conf. Comput. Vis. Pattern Recognit.},
  2016, pp. 770--778.

\bibitem{huang2017densely}
G.~Huang, Z.~Liu, L.~Van Der~Maaten, and K.~Q. Weinberger, ``Densely connected
  convolutional networks,'' in \emph{Proc. IEEE Conf. Comput. Vis. Pattern
  Recognit.}, 2017, pp. 4700--4708.

\bibitem{belongie2002shape}
S.~Belongie, J.~Malik, and J.~Puzicha, ``Shape matching and object recognition
  using shape contexts,'' \emph{IEEE Trans. Pattern Anal. Mach. Intell.},
  vol.~24, no.~4, pp. 509--522, 2002.

\bibitem{ramachandran2019stand}
P.~Ramachandran, N.~Parmar, A.~Vaswani, I.~Bello, A.~Levskaya, and J.~Shlens,
  ``Stand-alone self-attention in vision models,'' in \emph{Proc. Adv. Neural
  Inf. Process. Syst.}, 2019, pp. 68--80.

\bibitem{armeni20163d}
I.~Armeni \emph{et~al.}, ``3{D} semantic parsing of large-scale indoor
  spaces,'' in \emph{Proc. IEEE Conf. Comput. Vis. Pattern Recognit.}, 2016,
  pp. 1534--1543.

\bibitem{he2021masked}
\BIBentryALTinterwordspacing
K.~He \emph{et~al.}, ``Masked autoencoders are scalable vision learners,''
  \emph{arXiv:2111.06377}, 2021. [Online]. Available:
  \url{http://arxiv.org/abs/2111.06377}
\BIBentrySTDinterwordspacing

\bibitem{wu20153d}
Z.~Wu, S.~Song, A.~Khosla, F.~Yu, L.~Zhang, X.~Tang, and J.~Xiao, ``3{D}
  shapenets: A deep representation for volumetric shapes,'' in \emph{Proc. IEEE
  Conf. Comput. Vis. Pattern Recognit.}, 2015, pp. 1912--1920.

\bibitem{qi2019deep}
C.~R. Qi, O.~Litany, K.~He, and L.~J. Guibas, ``Deep hough voting for 3{D}
  object detection in point clouds,'' in \emph{Proc. IEEE Int. Conf. Comput.
  Vis.}, 2019, pp. 9277--9286.

\bibitem{xie2021vote}
Q.~Xie, Y.-K. Lai, J.~Wu, Z.~Wang, Y.~Zhang, K.~Xu, and J.~Wang, ``Vote-based
  3{D} object detection with context modeling and {SOB-3DNMS},'' \emph{Int. J.
  Comput. Vis.}, vol. 129, no.~6, pp. 1857--1874, 2021.

\bibitem{bai2021pointdsc}
X.~Bai \emph{et~al.}, ``{TransFusion}: robust lidar-camera fusion for 3{D}
  object detection with transformers,'' \emph{Proc. IEEE Conf. Comput. Vis.
  Pattern Recognit.}, pp. 1090--1099, 2022.

\bibitem{qi2020p2b}
H.~Qi, C.~Feng, Z.~Cao, F.~Zhao, and Y.~Xiao, ``P2{B}: Point-to-box network for
  3{D} object tracking in point clouds,'' in \emph{Proc. IEEE Conf. Comput.
  Vis. Pattern Recognit.}, 2020, pp. 6329--6338.

\bibitem{geiger2012we}
A.~Geiger, P.~Lenz, and R.~Urtasun, ``Are we ready for autonomous driving? the
  {KITTI} vision benchmark suite,'' in \emph{Proc. IEEE Conf. Comput. Vis.
  Pattern Recognit.}, 2012, pp. 3354--3361.

\bibitem{zeng20173dmatch}
A.~Zeng, S.~Song, M.~Nie{\ss}ner, M.~Fisher, J.~Xiao, and T.~Funkhouser,
  ``3{DM}atch: Learning local geometric descriptors from {RGB-D}
  reconstructions,'' in \emph{Proc. IEEE Conf. Comput. Vis. Pattern Recognit.},
  2017, pp. 1802--1811.

\bibitem{huang2021predator}
S.~Huang, Z.~Gojcic, M.~Usvyatsov, A.~Wieser, and K.~Schindler, ``{PREDATOR}:
  Registration of 3{D} point clouds with low overlap,'' in \emph{Proc. IEEE
  Conf. Comput. Vis. Pattern Recognit.}, 2021, pp. 4267--4276.

\bibitem{li2010action}
W.~Li, Z.~Zhang, and Z.~Liu, ``Action recognition based on a bag of 3{D}
  points,'' in \emph{Proc. IEEE Conf. Comput. Vis. Pattern Recognit.
  Workshops}, 2010, pp. 9--14.

\bibitem{shahroudy2016ntu}
A.~Shahroudy, J.~Liu, T.-T. Ng, and G.~Wang, ``{NTU RGB+D}: A large scale
  dataset for 3{D} human activity analysis,'' in \emph{Proc. IEEE Conf. Comput.
  Vis. Pattern Recognit.}, 2016, pp. 1010--1019.

\bibitem{liu2019ntu}
J.~Liu, A.~Shahroudy, M.~Perez, G.~Wang, L.-Y. Duan, and A.~C. Kot, ``{NTU
  RGB+D} 120: A large-scale benchmark for 3{D} human activity understanding,''
  \emph{IEEE Trans. Pattern Anal. Mach. Intell.}, vol.~42, no.~10, pp.
  2684--2701, 2019.

\bibitem{huang20223dpctn}
S.~Huang, Z.~Yang, Y.~Shi, J.~Tan, H.~Li, and Y.~Cheng, ``{3DPCTN}: Two 3{D}
  local-object point-cloud-completion transformer networks based on
  self-attention and multi-resolution,'' \emph{Electronics}, vol.~11, no.~9, p.
  1351, 2022.

\bibitem{wen2022pmp}
X.~Wen \emph{et~al.}, ``{PMP-N}et++: Point cloud completion by
  transformer-enhanced multi-step point moving paths,'' \emph{IEEE Trans.
  Pattern Anal. Mach. Intell.}, 2022.

\bibitem{Wei-AGConv2022}
\BIBentryALTinterwordspacing
M.~Wei \emph{et~al.}, ``{AGC}onv: Adaptive graph convolution on 3{D} point
  clouds,'' \emph{arXiv:2206.04665}, 2022. [Online]. Available:
  \url{http://arxiv.org/abs/2206.04665}
\BIBentrySTDinterwordspacing

\bibitem{yang2017foldingnet}
\BIBentryALTinterwordspacing
Y.~Yang, C.~Feng, Y.~Shen, and D.~Tian, ``Folding{N}et: Interpretable
  unsupervised learning on 3{D} point clouds,'' \emph{arXiv:1712.07262}, 2017.
  [Online]. Available: \url{http://arxiv.org/abs/1712.07262}
\BIBentrySTDinterwordspacing

\bibitem{yuan2018pcn}
W.~Yuan, T.~Khot, D.~Held, C.~Mertz, and M.~Hebert, ``P{CN}: Point completion
  network,'' in \emph{3DV}, 2018, pp. 728--737.

\bibitem{cciccek20163d}
{\"O}.~{\c{C}}i{\c{c}}ek, A.~Abdulkadir, S.~S. Lienkamp, T.~Brox, and
  O.~Ronneberger, ``3{D U-N}et: learning dense volumetric segmentation from
  sparse annotation,'' in \emph{Proc. Int. Conf. Med. Image Comput.
  Comput.-Assisted Intervention}, 2016, pp. 424--432.

\bibitem{zhao2019pointweb}
H.~Zhao, L.~Jiang, C.-W. Fu, and J.~Jia, ``Point{W}eb: Enhancing local
  neighborhood features for point cloud processing,'' in \emph{Proc. IEEE Conf.
  Comput. Vis. Pattern Recognit.}, 2019, pp. 5565--5573.

\bibitem{xu2018spidercnn}
Y.~Xu, T.~Fan, M.~Xu, L.~Zeng, and Y.~Qiao, ``Spider{CNN}: Deep learning on
  point sets with parameterized convolutional filters,'' in \emph{Eur. Conf.
  Comput. Vis.}, 2018, pp. 87--102.

\bibitem{li2018pointcnn}
Y.~Li, R.~Bu, M.~Sun, W.~Wu, X.~Di, and B.~Chen, ``Point{CNN}: Convolution on
  {X}-transformed points,'' in \emph{Proc. Adv. Neural Inf. Process. Syst.},
  2018, pp. 828--838.

\bibitem{wu2019pointconv}
W.~Wu, Z.~Qi, and L.~Fuxin, ``Point{C}onv: Deep convolutional networks on 3{D}
  point clouds,'' in \emph{Proc. IEEE Conf. Comput. Vis. Pattern Recognit.},
  2019, pp. 9621--9630.

\bibitem{lin2020fpconv}
Y.~Lin \emph{et~al.}, ``F{PC}onv: Learning local flattening for point
  convolution,'' in \emph{Proc. IEEE Conf. Comput. Vis. Pattern Recognit.},
  2020, pp. 4293--4302.

\bibitem{liu2019point2sequence}
X.~Liu, Z.~Han, Y.-S. Liu, and M.~Zwicker, ``Point2{S}equence: Learning the
  shape representation of 3{D} point clouds with an attention-based sequence to
  sequence network,'' in \emph{Proc. AAAI Conf. Artif. Intell.}, vol.~33,
  no.~01, 2019, pp. 8778--8785.

\bibitem{mao2019interpolated}
J.~Mao, X.~Wang, and H.~Li, ``Interpolated convolutional networks for 3{D}
  point cloud understanding,'' in \emph{Proc. IEEE Int. Conf. Comput. Vis.},
  2019, pp. 1578--1587.

\bibitem{zhang2019shellnet}
Z.~Zhang, B.-S. Hua, and S.-K. Yeung, ``Shell{N}et: Efficient point cloud
  convolutional neural networks using concentric shells statistics,'' in
  \emph{Proc. IEEE Int. Conf. Comput. Vis.}, 2019, pp. 1607--1616.

\bibitem{lee2021regularization}
D.~Lee \emph{et~al.}, ``Regularization strategy for point cloud via rigidly
  mixed sample,'' in \emph{Proc. IEEE Conf. Comput. Vis. Pattern Recognit.},
  2021, pp. 15\,900--15\,909.

\bibitem{xu2021paconv}
M.~Xu, R.~Ding, H.~Zhao, and X.~Qi, ``P{AC}onv: Position adaptive convolution
  with dynamic kernel assembling on point clouds,'' in \emph{Proc. IEEE Conf.
  Comput. Vis. Pattern Recognit.}, 2021, pp. 3173--3182.

\bibitem{ran2021learning}
H.~Ran, W.~Zhuo, J.~Liu, and L.~Lu, ``Learning inner-group relations on point
  clouds,'' in \emph{Proc. IEEE Conf. Comput. Vis. Pattern Recognit.}, 2021,
  pp. 15\,477--15\,487.

\bibitem{xiang2021walk}
T.~Xiang, C.~Zhang, Y.~Song, J.~Yu, and W.~Cai, ``Walk in the cloud: Learning
  curves for point clouds shape analysis,'' in \emph{Proc. IEEE Int. Conf.
  Comput. Vis.}, 2021, pp. 915--924.

\bibitem{ma2022rethinking}
\BIBentryALTinterwordspacing
X.~Ma, C.~Qin, H.~You, H.~Ran, and Y.~Fu, ``Rethinking network design and local
  geometry in point cloud: A simple residual {MLP} framework,''
  \emph{arXiv:2202.07123}, 2022. [Online]. Available:
  \url{http://arxiv.org/abs/2202.07123}
\BIBentrySTDinterwordspacing

\bibitem{yi2016scalable}
L.~Yi \emph{et~al.}, ``A scalable active framework for region annotation in
  3{D} shape collections,'' \emph{ACM Trans. Graph.}, vol.~35, no.~6, pp.
  1--12, 2016.

\bibitem{xia2022vision}
\BIBentryALTinterwordspacing
Z.~Xia, X.~Pan, S.~Song, L.~E. Li, and G.~Huang, ``Vision transformer with
  deformable attention,'' \emph{arXiv:2201.00520}, 2022. [Online]. Available:
  \url{http://arxiv.org/abs/2201.00520}
\BIBentrySTDinterwordspacing

\bibitem{zhu2021learning}
L.~Zhu \emph{et~al.}, ``Learning the superpixel in a non-iterative and lifelong
  manner,'' in \emph{Proc. IEEE Conf. Comput. Vis. Pattern Recognit.}, 2021,
  pp. 1225--1234.

\bibitem{hui2021superpoint}
L.~Hui, J.~Yuan, M.~Cheng, J.~Xie, X.~Zhang, and J.~Yang, ``Superpoint network
  for point cloud oversegmentation,'' in \emph{Proc. IEEE Int. Conf. Comput.
  Vis.}, 2021, pp. 5510--5519.

\bibitem{goodfellow2014generative}
I.~Goodfellow \emph{et~al.}, ``Generative adversarial nets,'' in \emph{Proc.
  Adv. Neural Inf. Process. Syst.}, 2014, pp. 2672--2680.

\bibitem{gadelha2018multiresolution}
M.~Gadelha, R.~Wang, and S.~Maji, ``Multiresolution tree networks for 3{D}
  point cloud processing,'' in \emph{Proc. Eur. Conf. Comput. Vis.}, 2018, pp.
  103--118.

\bibitem{yang2017towards}
B.~Yang, X.~Fu, N.~D. Sidiropoulos, and M.~Hong, ``Towards k-means-friendly
  spaces: Simultaneous deep learning and clustering,'' in \emph{Proc. Int.
  Conf. Mach. Learn.}, 2017, pp. 3861--3870.

\bibitem{achlioptas2018learning}
P.~Achlioptas, O.~Diamanti, I.~Mitliagkas, and L.~Guibas, ``Learning
  representations and generative models for 3{D} point clouds,'' in \emph{Proc.
  Int. Conf. Mach. Learn.}, 2018, pp. 40--49.

\end{thebibliography}
